\documentclass[onecolumn]{fairmeta}

\usepackage[most]{tcolorbox}

\usepackage{amsmath}
\usepackage{amssymb} \usepackage{bm}
\usepackage[utf8]{inputenc} 
\usepackage[T1]{fontenc}    
\usepackage{hyperref}       
\usepackage{url}            
\usepackage{booktabs}       
\usepackage{amsfonts}       
\usepackage{nicefrac}       
\usepackage{microtype}      
\usepackage{xcolor}         
\usepackage{graphicx}
\usepackage{float}
\newcommand{\m}[1]{{$\mathbf{\Delta}$}}
\newcommand{\p}[1]{}

\usepackage{wrapfig}
\usepackage{enumitem}
\usepackage{fontawesome5}

\title{\LARGE $\mathbf{\Delta}$ Flow Matching for Robot Foundation Models}

\author[1,*]{Kuangji Zuo}
\author[1,*]{\href{https://morpheus-an.github.io/}{\textcolor{black}{Tuo An}}}
\author[1]{Xinying Guo}
\author[1]{\href{https://reagan1311.github.io/}{\textcolor{black}{Gen Li}}}
\author[1]{Mengfei Zhao}
\author[1]{Jiaqi Bai}
\author[1]{Bofan Lyu}
\author[1]{Zhuoyuan Yu}
\author[1]{Jinghan Yang}
\author[1]{Jiayi Li}
\author[1,\dagger, \ddag]{\href{https://jiajindou.github.io/}{\textcolor{black}{Jindou Jia}}}

\author[1,\dagger]{\href{https://marsyang.site/}{\textcolor{black}{Jianfei Yang}}}

\affiliation[1]{MARS Lab, Nanyang Technological University}

\contribution[*]{Equal contribution}
\contribution[\ddag]{Project Lead}
\contribution[\dagger]{Corresponding authors}

\correspondence{\email{jindou.jia@ntu.edu.sg}, \email{jianfei.yang@ntu.edu.sg}.}
\metadata[Project page]{\url{https://zuo-kuangji.github.io/delta-flow/}}

\abstract{

Flow matching has emerged as a common generative recipe for current robot foundation models. Vision-language-action models (VLAs) generate action chunks, while world-action models (WAMs) additionally generate future visual states. Yet both start generation from a fixed \textit{Gaussian} prior, despite the strong temporal continuity between past and future states and actions. This process is costly to train and requires iterative sampling at deployment, especially for WAMs with an additional visual branch. Exploiting this temporal continuity, we introduce \textbf{\m{}-flow}, which constructs the source from recent history with mild noise rather than from a fixed \textit{Gaussian} prior. This moves the source closer to the target, reducing the transport to a small residual \m{} that can be learned efficiently and enables single-step inference. We integrate \m{}-flow into the VLA policy $\pi_{0.5}$ and the WAM policy including ImageWAM and Fast-WAM. On LIBERO, ImageWAM reaches $90\%$ success in one epoch rather than six, and Fast-WAM reaches $95\%$ in two epochs rather than eight. With LoRA fine-tuning, $\pi_{0.5}$ exceeds its baseline's five-epoch performance after one epoch. Single-step inference gives $1.54\times$, $2.49\times$, and $4.45\times$ speedups on $\pi_{0.5}$, ImageWAM, and Fast-WAM, respectively. It also preserves action smoothness, achieving a $4.5\times$ reduction in action jerk compared with one-step \textit{Gaussian} sampling. \m{}-flow reaches state-of-the-art generalization performance on LIBERO-Plus. Remarkably, action-history sources can also be applied to pretrained WAM policies without retraining. Replacing the \textit{Gaussian} action source at inference largely preserves task success with a single sampling step and can slightly improve it. We further study how action representation and the use of visual information in action learning affect history-to-future prediction in \m{}-flow. Together, these results show that \m{}-flow is \textit{simple yet effective}, substantially improving the efficiency and generalization of current robot foundation models.

}

\begin{document}

\maketitle

\section{Introduction} 

\begin{figure*}[t]
    \centering
    \includegraphics[width=\textwidth]{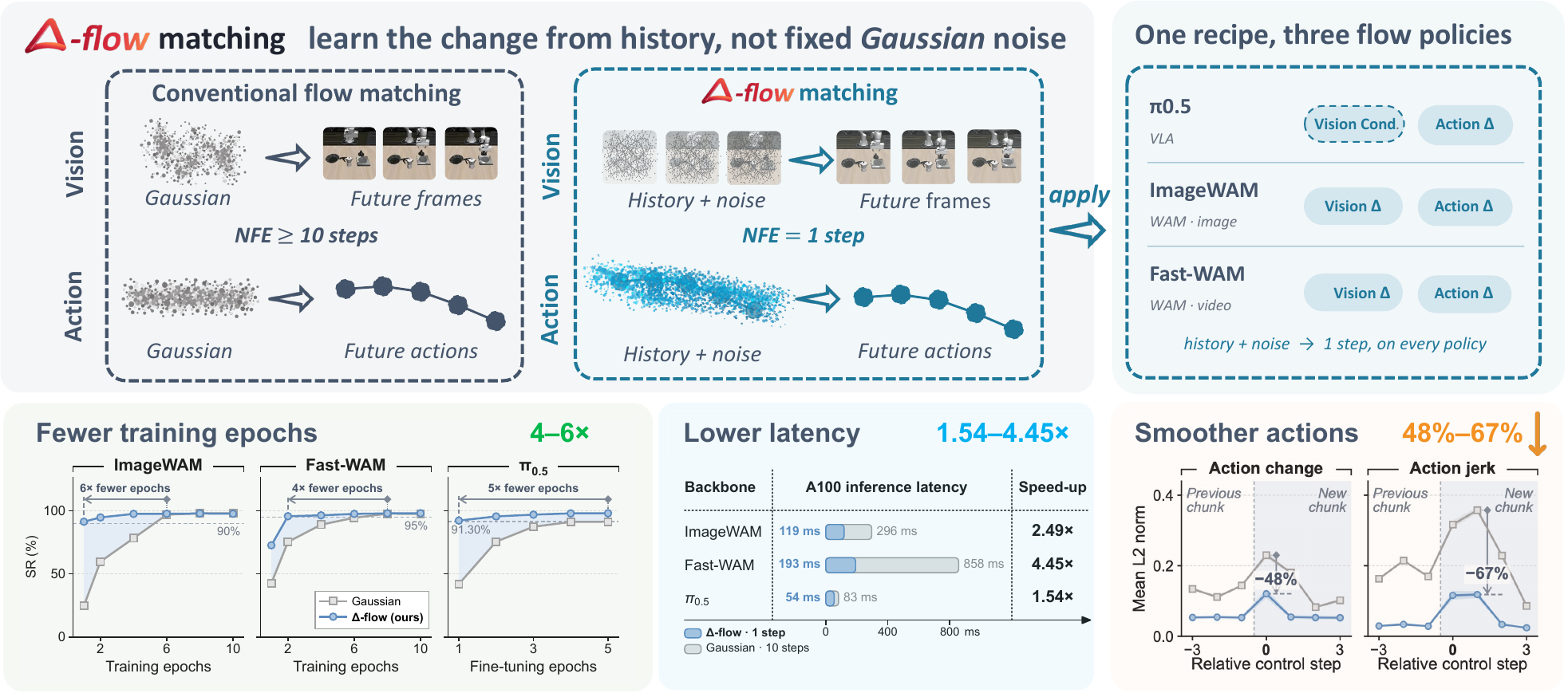}
    \caption{\m{}-flow matching learns past-to-future changes for efficient robot foundation models. Unlike \textit{Gaussian}-source flow matching, which reconstructs future visual states and actions from noise through iterative sampling, \m{}-flow constructs each source from its corresponding recent history with mild \textit{Gaussian} perturbation and learns only the residual change \m{}. This shorter transport substantially improves training efficiency, enables fast single-step inference, strengthens generalization under distribution shift, and preserves smooth action generation across chunk boundaries.}
    \label{fig:teaser}
\end{figure*}

Flow matching has become a standard recipe for continuous action generation in robot policies~\citep{lipman2023flow}. In vision-language-action models (VLAs), an action chunk is produced by transporting a sample from a \textit{Gaussian} prior to the action distribution, conditioned on the current observation~\citep{black2024pi0,intelligence2025pi05}. More recently, world-action models (WAMs) extend this generative formulation beyond actions to future visual representations, jointly modeling how the scene evolves and how the robot acts~\citep{dreamzero,fastwam,imagewam,hou2026world}. In both families, generation begins from \textit{Gaussian} noise.

However, this \textit{Gaussian}-source formulation is expensive, both to train and to deploy~\citep{hou2026world,WM_Survey}. Even for action-only VLAs, flow-based action generation converges slowly during training~\citep{intelligence2025pi05}, and at deployment it requires a large number of function evaluations (NFE) (i.e., inference steps), which directly increases control latency. WAMs inherit both costs and amplify them. During training, WAMs learn transports from pure \textit{Gaussian} noise to both future actions and high-dimensional future visual states. At deployment, WAMs that jointly sample visual states and actions require computation for both modalities at each sampling step. Together, these training and inference costs pose challenges for deploying flow-based policies on real robots.

Most existing work targets the latter, reducing inference cost along three lines. Efficient samplers shorten the sampling trajectory or distill iterative generation into a single prediction~\citep{lu2022dpmsolver,song2023consistency,prasad2024consistencypolicy,zhang2024flowpolicy}, yet they are developed primarily for action generation, and whether they remain effective on the visual branch of WAMs is still an open question. Some warm-start methods instead use separately trained predictors to provide informative initializations, followed by a few sampling steps for refinement~\citep{li2026step,warmstarts}. Other models like Fast-WAM~\citep{fastwam} and ImageWAM~\citep{imagewam} retain future prediction as a training objective but skip it at inference, so that its knowledge reaches the action branch only implicitly through the shared representation. These models thus trade explicit future conditioning for inference speed. Our experiments show that removing explicit future prediction at inference degrades performance under distribution shift. More fundamentally, these methods shorten, approximate, or partially bypass the generative transport to reduce inference cost, rather than reducing the cost of learning the transport itself.

Flow matching, in essence, learns a transport between two distributions~\citep{lipman2023flow,jia2026marspolicy,bai2026flash}. Since both actions and visual signals are naturally continuous, and their executed histories are directly available from feedback measurements, \textbf{why not take these histories as the source of the flow?} Using history as the source can simplify the mapping learned during training and shorten the transport path at inference. Several works in image and video generation indeed construct their sources from existing images or adjacent frames and learn transports to target images or future frames~\citep{pmlr-v202-liu23ai,Davtyan_2023_ICCV,Liu_2025_ICCV}. In robot control, action-to-action (A2A) flow matching uses histories of executed actions as the source distribution and learns a transport to future action chunks~\citep{a2a2026}. 
Each line, however, treats a single modality in isolation. Under the WAM paradigm, and especially at the scale of growing video and robot data, jointly training these two mechanisms, and identifying the design choices that make them work together, remains an open question.

In this work, we propose \textbf{\m{}-flow matching} for robot foundation models, termed \textbf{\m{}-flow}, where the symbol \m{} denotes the difference between the history and the future of a visual state or action. Rather than generating the future from fixed \textit{Gaussian} noise, \m{}-flow directly learns how the recent visual and action history evolves into its future, leaving the flow networks responsible only for residuals. In VLAs, this replaces the \textit{Gaussian} source of the action branch; in WAMs, it replaces the source of both branches. The future visual representations and actions of WAMs are modeled jointly, with action generation conditioned on the predicted visual future, and both produced in a single flow step at inference. This design preserves both action continuity and state–action coupling, which are identified as important factors for successful trajectories~\citep{yang2026tri}. Because \m{}-flow alters only the source of the flow, it integrates naturally into existing policies, including WAMs such as ImageWAM~\citep{imagewam} and Fast-WAM~\citep{fastwam} and flow-based VLAs such as $\pi_{0.5}$~\citep{intelligence2025pi05}, without network architectural modification or extra data.

Figure~\ref{fig:teaser} summarizes the formulation and its training and inference benefits. Experiments show that \m{}-flow is \textit{simple yet effective}. 
We evaluate \m{}-flow across robot foundation policies on LIBERO~\citep{liu2023libero} and LIBERO-Plus~\citep{fei2026liberoplus}.
 During training on LIBERO, ImageWAM reaches $90\%$ success in one rather than six epochs, while Fast-WAM reaches $95\%$ in two rather than eight. On $\pi_{0.5}$ with LoRA fine-tuning, one epoch exceeds the baseline's five-epoch success, giving a $5\times$ reduction in training epochs to that performance level. At inference, \m{}-flow accelerates all three backbones through single-step generation. On Fast-WAM, it jointly predicts future visual representations and actions $4.45\times$ faster than ten-step \textit{Gaussian} sparse-joint inference while improving LIBERO-Plus success. \m{}-flow also maintains smooth actions at one step: on ImageWAM, it achieves a $4.5\times$ reduction in action jerk compared with one-step \textit{Gaussian} sampling, with smoothness comparable to ten-step \textit{Gaussian} sampling. On LIBERO-Plus, it further achieves state-of-the-art performance without large robot training data, reaching $88.55\%$ success across the seven perturbation categories. \textbf{Most strikingly}, replacing the \textit{Gaussian} action source with recent executed actions enables single-step sampling in pretrained ImageWAM and Fast-WAM without retraining. This adaptation largely preserves task success and can slightly improve it.

Beyond the main results, we systematically examine how \m{}-based flows interact with existing policy and WAM designs, and several findings are counterintuitive. First, action-space parameterization barely changes standard LIBERO success but shifts LIBERO-Plus success by over $14$ points, so parameterizations that appear equivalent in distribution can differ sharply under perturbation (Figure~\ref{fig:iw-action-space}). Second, predicting fewer future frames over the same horizon improves control, gaining $5.19$ points on LIBERO-Plus over a denser schedule (Figure~\ref{fig:fastwam-variants}). Third, as shown in Figure~\ref{fig:iw_visual_stopgrad}, stop-gradient improves future-image PSNR and LIBERO-Plus success. The action branch still reads visual features, but its gradients do not update the world model. These results indicate that feature sharing and gradient coupling should be treated as separate design decisions. Finally, once the source is past-informed and visual supervision is retained, supervising only the one-step endpoint of the action branch nearly matches the full flow-matching objective, albeit with slower convergence (Figure~\ref{fig:iw-consistency-only}). We believe these findings are useful beyond the specific setting studied here.

\section{Related Work}

\subsection{Robot foundation models}
\subsubsection{Vision-language-action model} VLAs map visual observations and language instructions to robot actions~\citep{zitkovich2023rt2,kim2025openvla}. $\pi_0$~\citep{black2024pi0} introduces a flow-matching action expert for continuous action generation, while $\pi_{0.5}$~\citep{intelligence2025pi05} further improves open-world generalization through heterogeneous co-training. FAST~\citep{pertsch2025fast} compresses action sequences to reduce autoregressive decoding steps, while FLOWER~\citep{reuss2025flower} reduces action-expert parameters by sharing adaptive normalization across transformer layers. In flow-based VLA policies, generation starts from \textit{Gaussian} noise and iteratively transports it into a future action chunk conditioned on the observation~\citep{black2024pi0}. Such policies can be costly to train, and repeated action-expert evaluations increase inference latency~\citep{pertsch2025fast,black2024pi0}. Our work instead constructs the source from recently executed actions and learns their transformation into future action chunks.

\subsubsection{World-action models} 
WAMs couple predictive world modeling with robot action generation. Joint WAMs generate future visual representations and actions together~\citep{dreamzero,an2026feedback}, while causal WAMs condition action prediction on predicted future visual states~\citep{lingbotva}. To reduce inference costs, Fast-WAM~\citep{fastwam} retains video prediction during training but generates actions from current visual representations at test time. ImageWAM~\citep{imagewam} replaces video generation with image editing and conditions action generation on current-image representations. These methods mainly change how visual information supports action generation. However, both visual and action generation still start from \textit{Gaussian} noise. Our work instead introduces a unified history-based source formulation for visual and action generation, learning past-to-future changes in both modalities.

\subsection{Past-to-future generative modeling}

A few early efforts have begun to start generation from the past. In visual generation, several methods construct more informative sources from existing images. For example, I$^2$SB~\citep{pmlr-v202-liu23ai} learns a Schr\"odinger bridge from a source image distribution to a target image distribution, grounding image translation directly in the input image. Related video generation methods~\citep{Davtyan_2023_ICCV,Liu_2025_ICCV} exploit relationships between adjacent frames to learn the generation of future frames from observed ones. In robot control, action-to-action (A2A) flow matching~\citep{a2a2026} uses executed action histories as the source distribution and learns their transport to future action chunks. Warm-start methods~\citep{chen2024bridger,li2026step} instead use auxiliary policies or predictors to provide action initializations, followed by a small number of refinement steps. These studies explore informed sources separately for visual and action generation, but do not examine how both can be coordinated within the WAMs. Our work unifies past visual states and actions as generative sources, providing a common past-to-future formulation for future world and action generation.

\textbf{Relation to our prior A2A policy.}
\m{}-flow builds on our earlier A2A flow matching~\citep{a2a2026}, which transports executed action histories to future action chunks, and extends it in four respects. \textbf{(1)} A2A defines the transport in a learned latent space, whereas \m{}-flow operates directly in the action space, so it can be dropped into an existing policy without an additional encoder or decoder. \textbf{(2)} A2A establishes the idea for actions only; here we show that the same difference-based source is equally effective for visual generation, and that the two branches can be coupled within a single WAM. \textbf{(3)} A2A was evaluated on task-specific visuomotor policies, whereas we test the formulation at the scale of robot foundation models, covering both flow-based VLAs and WAMs. \textbf{(4)} We report a set of empirical studies absent from A2A, on action representation, world-action coupling, and knowledge insulation.

\section{Preliminaries}

\textbf{Flow matching.}
Flow matching generates a target $\mathbf{y}$ conditioned on context $c$ by integrating a velocity field $\mathbf{f}_{\theta}$ with learnable parameters $\theta$~\citep{lipman2023flow}.
For target-context pairs $(\mathbf{y},c)$ drawn from the training distribution $\mathcal{D}$, we sample \textit{Gaussian} noise $\bm{\epsilon}\sim\mathcal{N}(\mathbf{0},\mathbf{I})$ with the same dimensions as $\mathbf{y}$ and a flow time $\tau\sim\mathcal{U}(0,1)$, and construct the interpolated state $\mathbf{y}^{\tau}=(1-\tau)\mathbf{y}+\tau\bm{\epsilon}$.
The field is trained to predict the path velocity $\bm{\epsilon}-\mathbf{y}$ by minimizing
 \begin{equation} \mathcal{L}_{\mathrm{FM}}=\mathbb{E}_{(\mathbf{y},c)\sim\mathcal{D},\,\bm{\epsilon}\sim\mathcal{N}(\mathbf{0},\mathbf{I}),\,\tau\sim\mathcal{U}(0,1)}\left[\left\|\mathbf{f}_{\theta}(\mathbf{y}^{\tau},\tau,c)-(\bm{\epsilon}-\mathbf{y})\right\|_2^2\right]. \label{eq:standard-fm} \end{equation} At inference, generation starts from \textit{Gaussian} noise at $\tau=1$ and integrates the learned field toward $\tau=0$.

\textbf{Flow-based vision-language-action models.}
At control step $t$, a flow-based VLA predicts a chunk of $H$ future actions, $\mathbf{A}_t=[\mathbf{a}_t,\ldots,\mathbf{a}_{t+H-1}]$, from the available task context $c_t$. This context $c_t=(\mathbf{o}_{\leq t},\ell,\mathbf{q}_t)$ comprises visual observations $\mathbf{o}_{\leq t}$, the language instruction $\ell$, and any proprioceptive input $\mathbf{q}_t$.
Each action has dimension $d_a$, giving the chunk a flattened dimensionality of $D_a=Hd_a$.
 Flow-based VLAs use an action expert conditioned on a pretrained vision-language backbone to generate $\mathbf{A}_t$~\citep{black2024pi0,intelligence2025pi05}. 
During training, the demonstrated action chunk is interpolated with \textit{Gaussian} noise $\bm{\epsilon}_a\sim\mathcal{N}(\mathbf{0}_{D_a},\mathbf{I}_{D_a})$ to obtain $\mathbf{A}_t^\tau=(1-\tau)\mathbf{A}_t+\tau\bm{\epsilon}_a$, where $\tau$ denotes flow time rather than control time $t$.
Following Eq.~\eqref{eq:standard-fm}, the action expert $\mathbf{f}_{\theta_a}$, with learnable parameters $\theta_a$, predicts the velocity $\bm{\epsilon}_a-\mathbf{A}_t$ from $(\mathbf{A}_t^\tau,\tau,c_t)$.
 At inference, a chunk is initialized with \textit{Gaussian} noise and iteratively updated using the learned velocity field: 
\begin{equation}
\mathbf{A}_t^{\tau-\delta}=\mathbf{A}_t^\tau-\delta\mathbf{f}_{\theta_a}(\mathbf{A}_t^\tau,\tau,c_t),\qquad \delta=1/N.
\label{eq:vla-euler}
\end{equation}
Here, $\delta=1/N$ is the flow-time step size, and $N$ is the number of Euler steps, or equivalently the number of function evaluations (NFE) for the action field.
 Each update refines the entire action chunk as flow time decreases from $1$ to $0$.

\textbf{Flow-based world-action models.}
WAMs augment action prediction with future visual modeling.
A visual encoder $\mathbf{E}_v$ maps the future image or video frames to a latent target $\mathbf{Z}_t$.
The target $\mathbf{Z}_t$ has a flattened dimensionality of $D_v$.
 We take latent-space prediction as the example throughout;
models that generate pixels directly are recovered by letting $\mathbf{E}_v$ be
the identity. 
The flow-matching objective is applied to both targets, with $\mathbf{y}=\mathbf{A}_t$ for the action branch and $\mathbf{y}=\mathbf{Z}_t$ for the visual branch.
For the visual branch, $\mathbf{Z}_t^\tau=(1-\tau)\mathbf{Z}_t+\tau\bm{\epsilon}_v$, where $\bm{\epsilon}_v\sim\mathcal{N}(\mathbf{0}_{D_v},\mathbf{I}_{D_v})$. The visual field $\mathbf{f}_{\theta_v}$, with learnable parameters $\theta_v$, predicts $\bm{\epsilon}_v-\mathbf{Z}_t$ from $(\mathbf{Z}_t^\tau,\tau,c_t)$, while the action field $\mathbf{f}_{\theta_a}$ follows the formulation above.
 The training objective is
$\mathcal{L}_{\mathrm{WAM}}=\lambda_{v}\mathcal{L}_{\mathrm{FM}}^{v}
+\lambda_{a}\mathcal{L}_{\mathrm{FM}}^{a}$, where
$\mathcal{L}_{\mathrm{FM}}^{v}$ and $\mathcal{L}_{\mathrm{FM}}^{a}$ apply
Eq.~\eqref{eq:standard-fm} to the visual and action branches respectively, and
$\lambda_{v},\lambda_{a}\geq0$ balance the two terms. Visual and action tokens
can interact during joint generation~\citep{dreamzero}, or the action expert can
condition on intermediate visual features without requiring a decoded future
image~\citep{imagewam}. At inference, the generated variables are updated using
their predicted velocities, as in Eq.~\eqref{eq:vla-euler}, with
architecture-specific sampling budgets.

\begin{figure*}[t]
    \centering
    \includegraphics[width=\textwidth]{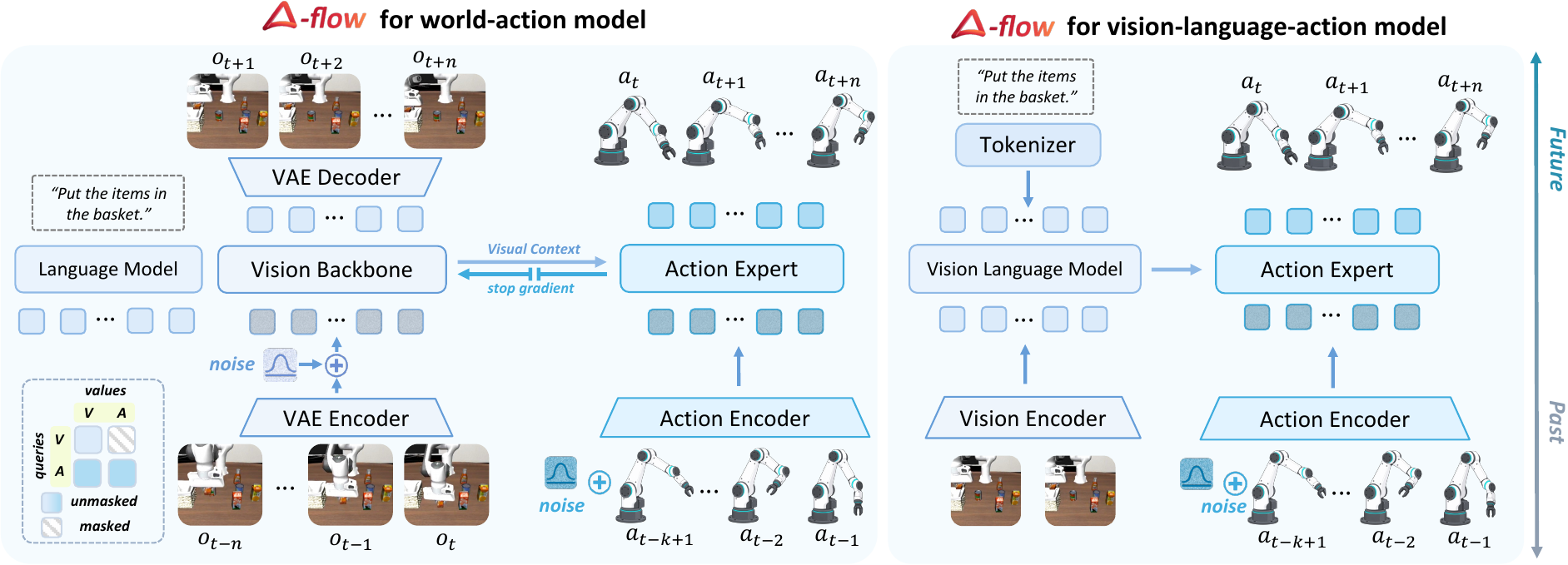}
    \caption{\textbf{\m{}-flow for robot foundation policies.}
    History-based sources combine past representations with \textit{Gaussian} noise, so the model learns past-to-future changes.
    \textbf{Left:} $\Delta$-flow for world-action models. The model learns both visual and action changes from history to future targets. The action expert uses intermediate visual representations, with stop-gradient blocking action-loss gradients from updating the visual branch during training.
    \textbf{Right:} $\Delta$-flow for vision-language-action models. The model learns action changes from executed history to future actions, using the original vision-language conditioning.
    At inference, a single Euler update transforms each history-based source into its future prediction.}
    \label{fig:delta0_framework}
\end{figure*}

\section{Methods}
\label{sec:method}\m{}-flow matching learns past-to-future changes for robot foundation policies. Figure~\ref{fig:teaser} illustrates its key idea: the source is built from historical representations with \textit{Gaussian} perturbations, so that the flow only has to carry the change from historical actions and visual representations to their corresponding future targets.

Let $m\in\{a,v\}$ denote the action or visual modality. We sample paired
historical and future representations $(\mathbf{h}_m,\mathbf{y}_m,c_t)\sim\mathcal{D}$
with $\mathbf{h}_m,\mathbf{y}_m\in\mathbb{R}^{D_m}$. The action target is
$\mathbf{y}_a=\mathbf{A}_t$ and the visual target is $\mathbf{y}_v=\mathbf{Z}_t$.
For actions, $\mathbf{h}_a=\mathbf{A}_t^{\mathrm{hist}}=[\mathbf{a}_{t-H},\ldots,\mathbf{a}_{t-1}]$
collects the $H$ most recently executed actions. For vision,
$\mathbf{h}_v=\mathbf{P}_v(\mathbf{E}_v(\mathbf{o}_{\leq t}))$ represents the
observed visual context, where the backbone-specific operator $\mathbf{P}_v$
arranges the encoded observations in the latent layout of the future image or
video target.

Instead of a fixed \textit{Gaussian} prior, we define the source of modality $m$
as its perturbed history
\begin{equation}
    \mathbf{s}_m=\alpha_m\mathbf{h}_m+\sigma_m\mathbf{b}_m\odot\bm{\epsilon}_m,
    \qquad
    \bm{\epsilon}_m\sim\mathcal{N}(\mathbf{0}_{D_m},\mathbf{I}_{D_m}),
    \label{eq:difference-source}
\end{equation}
where $\alpha_m$ and $\sigma_m$ control the history and noise scales,
$\mathbf{b}_m$ masks noise across dimensions, and $\odot$ denotes elementwise
multiplication. Unlike A2A flow matching~\citep{a2a2026}, which transports the clean history itself, we perturb it with a certain degree of noise. A deterministic source would map each history to a single target, sacrificing the multimodality of the target distribution and robustness~\citep{jia2026marspolicy}.

Interpolating the target with this source,
$\mathbf{y}_m^\tau=(1-\tau)\mathbf{y}_m+\tau\mathbf{s}_m$, the vector field is
trained to predict the velocity along the path:
\begin{equation}
    \mathcal{L}_{\mathrm{FM}}^m
    =\mathbb{E}\left[\left\|
    \mathbf{f}_{\theta_m}(\mathbf{y}_m^\tau,\tau,c_t^m)
    -\left(\mathbf{s}_m-\mathbf{y}_m\right)\right\|_2^2\right],
    \label{eq:difference-fm}
\end{equation}
where $c_t^m$ denotes the conditioning available to modality $m$, and the
expectation is over paired training samples, source noise, flow time
$\tau\sim\mathcal{U}(0,1)$, and any sampled visual conditioning described below.

To supervise single-step generation directly, we apply one Euler update from the
source at $\tau=1$ to $\tau=0$ and penalize its discrepancy from the target:
\begin{equation}
    \widehat{\mathbf{y}}_m
    =\mathbf{s}_m-\mathbf{f}_{\theta_m}(\mathbf{s}_m,1,c_t^m),
    \qquad
    \mathcal{L}_{\mathrm{EC}}^m
    =\mathbb{E}\left[\left\|\widehat{\mathbf{y}}_m-\mathbf{y}_m\right\|_1\right].
    \label{eq:endpoint-consistency}
\end{equation}
This expectation covers training samples, source noise, and any sampled visual
conditioning, but not flow time: the one-step prediction is always evaluated at
$\tau=1$. Flow matching thus supervises the vector field at sampled flow times,
whereas endpoint consistency supervises one-step generation. Action losses are
evaluated in the policy's normalized action space and visual losses in the WAM's
latent space.

Figure~\ref{fig:delta0_framework} illustrates how \m{}-flow is instantiated in VLAs and WAMs, which we describe in turn below.

\subsection{\m{}-Flow for Vision-Language-Action Models}

A VLA generates only the action chunk, so \m{}-flow applies to the action
modality alone. We set $\mathbf{h}_a=\mathbf{A}_t^{\mathrm{hist}}$,
$\mathbf{y}_a=\mathbf{A}_t$ and $c_t^a=c_t$, while retaining the
vision-language conditioning and the action-expert architecture. The training
objective combines the two terms of Eqs.~\eqref{eq:difference-fm}
and~\eqref{eq:endpoint-consistency} at $m=a$,
\begin{equation}
    \mathcal{L}_{\mathrm{VLA}}
    =\mathcal{L}_{\mathrm{FM}}^a+\lambda_{\mathrm{EC}}^a\mathcal{L}_{\mathrm{EC}}^a,
    \label{eq:delta-vla-objective}
\end{equation}
with $\lambda_{\mathrm{EC}}^a\geq0$ weighting action endpoint consistency. At
inference, a single Euler update produces the action chunk:
\begin{equation}
    \widehat{\mathbf{A}}_t
    =\mathbf{s}_a-\mathbf{f}_{\theta_a}(\mathbf{s}_a,1,c_t).
    \label{eq:delta-vla-inference}
\end{equation}
The source is constructed from executed actions, without an auxiliary history
predictor.

\subsection{\m{}-Flow for World-Action Models}

A WAM learns both visual and action changes, with paired targets drawn from the
same interaction window. The visual branch supplies intermediate
representations of current and future visual tokens to condition action
generation, and we stop gradients at this connection so that the action loss
does not update the visual branch through it:
\begin{equation}
    \mathbf{R}_t^{\tau_v}=\mathbf{R}_{\theta_v}(\mathbf{y}_v^{\tau_v},\tau_v,c_t),
    \qquad
    c_t^a=\left(c_t,\operatorname{sg}(\mathbf{R}_t^{\tau_v})\right),
    \label{eq:delta-zero-training-coupling}
\end{equation}
where $\operatorname{sg}$ denotes stop-gradient and $\tau_v$ is the visual flow
time specified by the backbone and training objective. The visual conditioning
used by each objective is detailed in the architecture instantiations below. The
two modalities are then trained jointly,
\begin{equation}
    \mathcal{L}_{\mathrm{WAM}}
    =\lambda_v\left(\mathcal{L}_{\mathrm{FM}}^v+\lambda_{\mathrm{EC}}^v\mathcal{L}_{\mathrm{EC}}^v\right)
    +\lambda_a\left(\mathcal{L}_{\mathrm{FM}}^a+\lambda_{\mathrm{EC}}^a\mathcal{L}_{\mathrm{EC}}^a\right),
    \label{eq:delta-zero-objective}
\end{equation}
where each term follows Eqs.~\eqref{eq:difference-fm}
and~\eqref{eq:endpoint-consistency} at $m=v$ and $m=a$, conditioned on $c_t$
and $c_t^a$ respectively. The nonnegative weights $\lambda_v,\lambda_a$ balance
visual and action supervision, while $\lambda_{\mathrm{EC}}^v,\lambda_{\mathrm{EC}}^a$
control endpoint consistency within each branch.

At inference, the visual network is evaluated at its source to obtain the
features used for action conditioning,
\begin{equation}
    \mathbf{R}_t^1=\mathbf{R}_{\theta_v}(\mathbf{s}_v,1,c_t),
    \label{eq:delta-zero-inference}
\end{equation}
and the visual endpoint $\widehat{\mathbf{Z}}_t$ and action chunk
$\widehat{\mathbf{A}}_t$ each take one Euler update as in
Eq.~\eqref{eq:endpoint-consistency}, the latter conditioned on
$(c_t,\operatorname{sg}(\mathbf{R}_t^1))$. The future visual endpoint therefore
never needs to be decoded.

\subsection{Architecture Instantiations}

Across all three policies, we construct the action source from the most recent executed actions, using a history window equal in length to the predicted action chunk. The actions are normalized using each policy's action statistics. We set $\alpha_a=1$ and $\sigma_a=0.5$ in Eq.~\eqref{eq:difference-source}, with the noise mask excluding the gripper and any padded dimensions.

\textbf{{ImageWAM Instantiation.}}
We instantiate \m{}-flow matching in ImageWAM, which combines the FLUX.2-klein-base-4B image-editing backbone with an action DiT~\citep{imagewam}. The visual history $\mathbf{h}_v$ is the encoded current image, arranged in the target latent layout, and the target $\mathbf{y}_v$ is the encoded image 16 frames into the future. We construct the visual source as $\mathbf{s}_v=(\mathbf{h}_v+\bm{\epsilon}_v)/\sqrt{2}$, corresponding to a 50\% image-latent prior. The action history contains the previous 16 executed actions. The action history $\mathbf{h}_a=\mathbf{A}_t^{\mathrm{hist}}$ and target $\mathbf{y}_a=\mathbf{A}_t$ are represented in the same normalized action space. The action expert reads layer-wise key--value features produced by the visual forward pass at the history-based source. These features contain intermediate representations of current and future visual tokens and are detached before conditioning the action expert. Thus, visual generation learns the current-to-future image change, while action generation learns the change from executed to future actions, with both branches accounting for the injected source noise. We set $\lambda_a=1$, $\lambda_v=0.5$, and $\lambda_{\mathrm{EC}}^a=\lambda_{\mathrm{EC}}^v=0.5$. ImageWAM training and evaluation settings are detailed in Appendix~\ref{app:imagewam-setup}.

\textbf{{Fast-WAM Instantiation.}}
We instantiate \m{}-flow matching in the joint-attention variant of Fast-WAM, which combines the Wan2.2-TI2V-5B video backbone with an action DiT~\citep{fastwam}. We use the sparse joint configuration (Fast-WAM-Joint-Sparse, also denoted Sparse8), which predicts four future frames at temporal stride $r=8$ over a 32-step horizon; its dense counterpart predicts eight frames at $r=4$ over the same horizon. We encode the observed frames at $\{t-32,t-24,t-16,t-8,t\}$ and use $\mathbf{P}_v$ to arrange their latents in the temporal layout of the future-video target $\mathbf{y}_v=\mathbf{Z}_t$, forming $\mathbf{h}_v$. The visual source is $\mathbf{s}_v=(\mathbf{h}_v+\bm{\epsilon}_v)/\sqrt{2}$, and the action history contains the previous 32 executed actions. The action branch uses the executed action history $\mathbf{h}_a=\mathbf{A}_t^{\mathrm{hist}}$ to predict $\mathbf{y}_a=\mathbf{A}_t$. Action tokens attend to intermediate representations of both current and future video tokens, with action-loss gradients stopped at this connection. Action flow matching uses visual features at sampled flow times, whereas action endpoint consistency uses features computed from the visual source. This instantiation uses future visual tokens and is distinct from the original Fast-WAM inference configuration that removes those tokens. We set $\lambda_a=\lambda_v=1$ and $\lambda_{\mathrm{EC}}^a=\lambda_{\mathrm{EC}}^v=0.5$. Fast-WAM training and evaluation settings are detailed in Appendix~\ref{app:fastwam-setup}.

\textbf{$\bm{\pi_{0.5}}$ Instantiation.} 
For $\pi_{0.5}$~\citep{intelligence2025pi05}, we retain the PaliGemma vision-language backbone and the Gemma-based action expert, and apply \m{}-flow matching to action generation. The history $\mathbf{h}_a=\mathbf{A}_t^{\mathrm{hist}}$ and target $\mathbf{y}_a=\mathbf{A}_t$ are represented in the checkpoint's quantile-normalized action space. We train with the action flow-matching and endpoint-consistency losses and use one Euler update at inference. No visual generation branch is introduced, as illustrated in Figure~\ref{fig:delta0_framework} (right). We use $\lambda_{\mathrm{EC}}^a=0.5$. The $\pi_{0.5}$ training and evaluation settings are detailed in Appendix~\ref{app:pi05-setup}.

\section{Evaluation}
\label{sec:experiments}

We evaluate how learning past-to-future changes affects training efficiency, generalization, and inference in robot foundation policies. We also test whether action-history sources can benefit pretrained WAM policies without updating their weights. Our evaluation addresses four questions: \textbf{(1)} Does \m{}-flow converge faster than its \textit{Gaussian}-source counterpart? \textbf{(2)} How does \m{}-flow affect generalization under distribution shift? \textbf{(3)} Can \m{}-flow reduce inference latency while maintaining action continuity? \textbf{(4)} Can action-history source initialization benefit a trained policy without retraining? 

\subsection{Training and Evaluation Setup}

We evaluate \m{}-flow on three prevalent robot foundation models, ImageWAM~\citep{imagewam}, Fast-WAM~\citep{fastwam}, and $\pi_{0.5}$~\citep{intelligence2025pi05}, comparing each with its original \textit{Gaussian}-source counterpart.
 In ImageWAM and Fast-WAM, \m{}-flow applies to both visual and action generation; in $\pi_{0.5}$, it applies only to action generation, with the vision-language conditioning unchanged.
 For Fast-WAM, we additionally include a \textit{Gaussian} sparse-joint baseline with the same Sparse8 future-frame layout, joint-attention architecture, and sampling budgets to assess the source design.
 Training-free experiments use frozen ImageWAM and Fast-WAM policies and change only the action-source initialization.

We evaluate task performance on the Spatial, Object, Goal, and Long suites of LIBERO~\citep{liu2023libero}, and assess generalization on LIBERO-Plus~\citep{fei2026liberoplus} without additional fine-tuning.
 The latter covers seven perturbation categories: camera
viewpoint, robot initialization, language instructions, lighting,
background textures, sensor noise, and object layout.
For our results, LIBERO Overall pools successful evaluations across the four suites.
Training progress is indexed by epochs or cumulative sample exposure
to account for differences in batch size. Detailed training and evaluation settings for ImageWAM, Fast-WAM, and $\pi_{0.5}$ are provided in Appendices~\ref{app:imagewam-setup}, \ref{app:fastwam-setup}, and \ref{app:pi05-setup}, respectively.

\begin{figure}[htbp]
\centering
\includegraphics[width=1.0\linewidth]{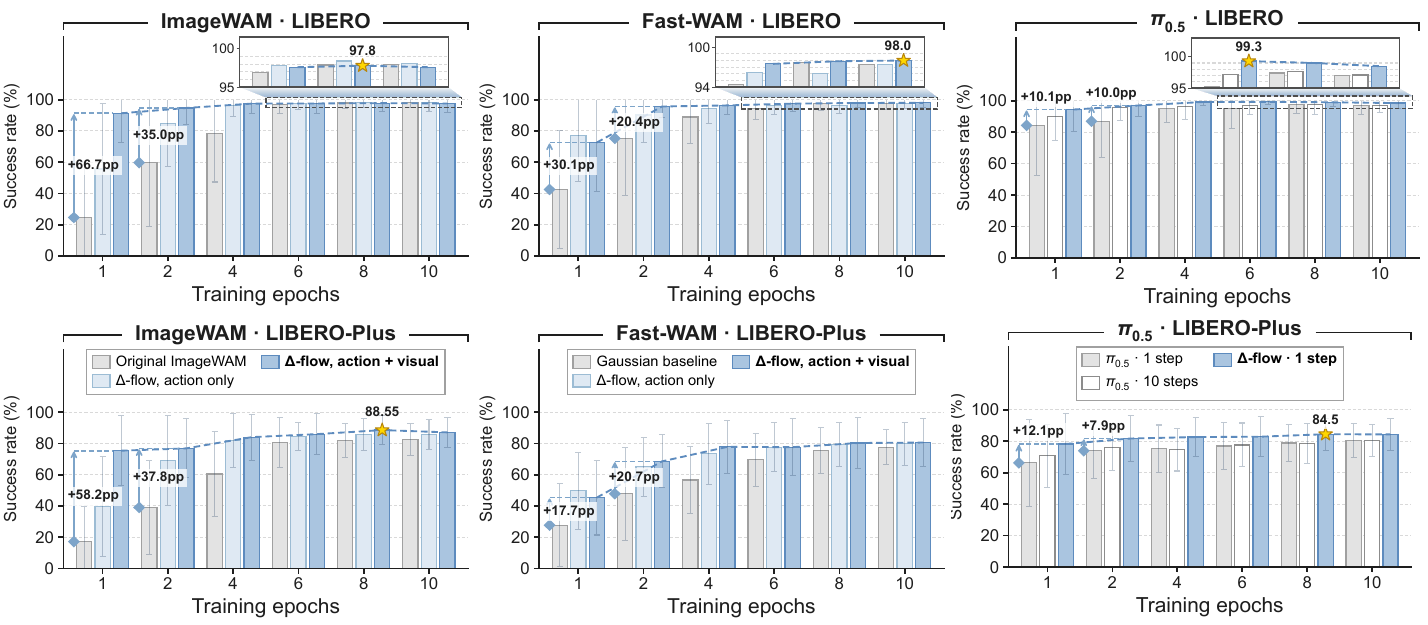}
\caption{\textbf{Training progress on LIBERO and LIBERO-Plus.}
Success rates are shown over training epochs for ImageWAM, Fast-WAM, and $\pi_{0.5}$.
The top panels show LIBERO and the bottom panels show LIBERO-Plus, evaluated without additional fine-tuning.
White bars denote the original models, gray bars denote the Gaussian sparse-joint Fast-WAM baseline, light-blue bars denote action-only $\Delta$-flow, and blue bars denote complete $\Delta$-flow.
The $\pi_{0.5}$ panels compare one-step and ten-step Gaussian sampling with one-step action $\Delta$-flow.
Dashed lines track $\Delta$-flow, and annotations report percentage-point differences.
Insets show later training epochs, and stars mark the highest evaluated success rate.
Error bars show one standard deviation across the 40 base tasks.}

\label{fig:training-efficiency_and_generalization}
\end{figure}

\subsection{Training Efficiency}

We first compare LIBERO success throughout training with the corresponding \textit{Gaussian}-source references. As shown in the top panels of Figure~\ref{fig:training-efficiency_and_generalization}, complete \m{}-flow on ImageWAM reaches $91.30\%$ after one epoch, compared with $24.60\%$ for the original model, a gain of $66.70$ percentage points. It first exceeds $90\%$ at epoch one, whereas the original first exceeds this threshold at the evaluated epoch-six checkpoint, giving a $6\times$ reduction in training epochs to this level. Action-only \m{}-flow reaches $90\%$ by epoch four. On Fast-WAM, \m{}-flow reaches $72.60\%$ after one epoch, compared with $42.55\%$ for the \textit{Gaussian} sparse-joint baseline. It reaches $95.65\%$ by epoch two, a level the baseline first exceeds at epoch eight, giving a $4\times$ reduction in training epochs. Moreover, applying \m{}-flow to actions alone already recovers most of this gain, and adding the visual source further accelerates learning. With the demonstration set held fixed, training progress is measured in epochs or cumulative samples processed.

\begin{figure}[htbp]
\centering
\includegraphics[width=0.9\linewidth]{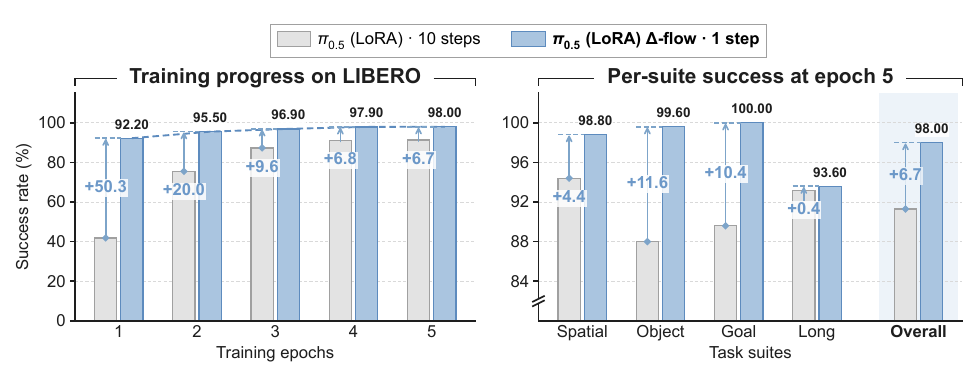}
\caption{\textbf{Training efficiency on $\pi_{0.5}$ with LoRA fine-tuning.} The left panel shows LIBERO success over five training epochs. The right panel shows success across the four LIBERO suites at epoch five. The curves and bars compare the ten-step Gaussian-source baseline with one-step action $\Delta$-flow. Annotations give percentage-point differences.}
\label{fig:pi05_lora_efficiency}
\end{figure}

As shown in the top-right panel of Figure~\ref{fig:training-efficiency_and_generalization}, action \m{}-flow on $\pi_{0.5}$ with full-parameter fine-tuning reaches at least $95\%$ LIBERO success by epoch two and peaks at $99.30\%$. All \m{}-flow models use a single Euler update at inference. The left panel of Figure~\ref{fig:pi05_lora_efficiency} shows that, with LoRA fine-tuning~\citep{hu2022lora}, action \m{}-flow reaches $92.20\%$ after one epoch, $95.50\%$ after two, and $98.00\%$ after five. Its first-epoch result already exceeds the ten-step baseline's $91.30\%$ after five epochs, giving a $5\times$ reduction in training epochs to exceed that performance level.

Across the WAM and VLA policies, \m{}-flow reaches high LIBERO success earlier in training. This advantage persists under LoRA fine-tuning on $\pi_{0.5}$, where one epoch of action \m{}-flow already exceeds the baseline's five-epoch performance. The reason is visible in what the network must fit. A \textit{Gaussian} source pairs each target with an unrelated noise sample, so two nearby sources can map to entirely different futures, forcing the velocity field to be steep, i.e., to have a large \textit{Lipschitz} constant~\citep{jia2026marspolicy}. A past-informed source instead pairs each target with its own history, leaving a small and smooth offset to be fit.

\setlength{\textfloatsep}{8pt plus 2pt minus 2pt}

\begin{table}[!t]
\centering
\caption{\textbf{Success rates on LIBERO and LIBERO-Plus.}
Success rates are reported in \%.
Emb.PT indicates additional embodied-data pretraining, excluding general-purpose vision, language, or video backbone pretraining.
Bold marks the highest value in each column within the VLA and WAM groups separately.
Literature results retain their original reporting protocols.
For models evaluated by us, each \m{}-flow variant and its original counterpart use the same hardware and matched evaluation settings.
Both Fast-WAM-joint rows use the sparse joint configuration with one-step sampling.}
\label{tab:main-results}

\begingroup
\fontsize{9}{11}\selectfont\rmfamily
\setlength{\tabcolsep}{2pt}
\renewcommand{\arraystretch}{1.12}

\textbf{(a) LIBERO}\par\vspace{3pt}
\begin{tabular*}{\linewidth}{@{\extracolsep{\fill}}l*{6}{c}@{}}
\toprule
Model & Emb.PT & Spatial & Object & Goal & Long & Overall \\
\midrule
\multicolumn{7}{@{}l@{}}{\textbf{VLA}} \\
OpenVLA~\citep{kim2025openvla}
    & $\checkmark$ & 84.7 & 88.4 & 79.2 & 53.7 & 76.5 \\
UniVLA~\citep{bu2025univla}
    & $\checkmark$ & 96.5 & 96.8 & 95.6 & 92.0 & 95.2 \\
$\pi_{0.5}$~\citep{intelligence2025pi05}
    & $\checkmark$ & 98.80 & 96.80 & 98.80 & 94.00 & 97.10 \\
$\pi_{0.5}$ (\m{}-flow)
    & $\checkmark$ & \textbf{99.60} & \textbf{100.00} & \textbf{99.60}
    & \textbf{96.80} & \textbf{99.00} \\
\midrule
\multicolumn{7}{@{}l@{}}{\textbf{WAM}} \\
LingBot-VA~\citep{lingbotva}
    & $\checkmark$ & 98.5 & 99.6 & 97.2 & \textbf{98.5} & 98.5 \\
Motus~\citep{bi2025motus}
    & $\checkmark$ & 96.8 & 99.8 & 96.6 & 97.6 & 97.7 \\
ABot-M0.5~\citep{chen2026abotm05}
    & $\checkmark$ & \textbf{100.0} & 99.8 & \textbf{99.4}
    & 98.4 & \textbf{99.4} \\
Cosmos Policy~\citep{kim2026cosmospolicy}
    & $\times$ & 98.1 & \textbf{100.0} & 98.2 & 97.6 & 98.5 \\
ImageWAM~\citep{imagewam}
    & $\times$ & 98.80 & 99.20 & 98.80 & 95.20 & 98.00 \\
ImageWAM (\m{}-flow)
    & $\times$ & 99.20 & 99.60 & 98.40 & 94.00 & 97.80 \\
Fast-WAM~\citep{fastwam}
    & $\times$ & 97.60 & \textbf{100.00} & 95.80 & 93.00 & 96.60 \\
Fast-WAM-joint (\textit{Gaussian})
    & $\times$ & 98.40 & 99.40 & 98.60 & 93.40 & 97.45 \\
Fast-WAM-joint (\m{}-flow)
    & $\times$ & 98.00 & \textbf{100.00} & 98.60 & 95.40 & 98.00 \\
\bottomrule
\end{tabular*}

\par\vspace{6pt}

\textbf{(b) LIBERO-Plus}\par\vspace{3pt}
\begin{tabular*}{\linewidth}{@{\extracolsep{\fill}}l*{9}{c}@{}}
\toprule
Model & Emb.PT & Camera & Robot & Language & Light
      & Background & Noise & Layout & Overall \\
\midrule
\multicolumn{10}{@{}l@{}}{\textbf{VLA}} \\
OpenVLA~\citep{fei2026liberoplus}
    & $\checkmark$ & 0.8 & 3.5 & 23.0 & 8.1
    & 34.8 & 15.2 & 28.5 & 15.6 \\
UniVLA~\citep{imagewam}
    & $\checkmark$ & 1.8 & 46.2 & 69.6 & 69.0
    & 81.0 & 21.2 & 31.9 & 42.9 \\
$\pi_{0.5}$~\citep{intelligence2025pi05}
    & $\checkmark$ & 66.04 & 73.61 & 81.39 & \textbf{97.37}
    & 91.36 & 76.70 & 86.75 & 80.70 \\
$\pi_{0.5}$ (\m{}-flow)
    & $\checkmark$ & \textbf{73.36} & \textbf{79.35} & \textbf{83.80} & 96.94
    & \textbf{97.30} & \textbf{81.26} & \textbf{87.34} & \textbf{84.53} \\
\midrule
\multicolumn{10}{@{}l@{}}{\textbf{WAM}} \\
LingBot-VA~\citep{lingbotva}
    & $\checkmark$ & 40.9 & 83.0 & 86.4 & 82.3
    & 53.1 & 64.4 & 76.2 & 69.5 \\
Motus~\citep{bi2025motus}
    & $\checkmark$ & 48.0 & \textbf{88.0} & 89.0 & 83.0
    & 78.0 & 57.0 & \textbf{89.0} & 76.0 \\
ABot-M0.5~\citep{chen2026abotm05}
    & $\checkmark$ & 70.5 & 87.4 & 88.6 & 94.0
    & 89.7 & 75.5 & 85.2 & 83.4 \\
Cosmos Policy~\citep{zhang2026wamrobustness}
    & $\times$ & 75.8 & 63.3 & 81.7 & 96.5
    & 88.9 & 92.7 & 82.2 & 82.2 \\
ImageWAM~\citep{imagewam}
    & $\times$ & \textbf{82.55} & 52.90 & 86.99 & 97.37
    & 90.24 & 92.07 & 81.31 & 82.49 \\
ImageWAM (\m{}-flow)
    & $\times$ & 82.24 & 73.94 & 93.62
    & \textbf{98.69} & \textbf{93.77} & \textbf{93.57} & 88.39 & \textbf{88.55} \\
Fast-WAM~\citep{fastwam}
    & $\times$ & 44.8 & 64.2 & 70.2 & 89.0
    & 57.3 & 75.0 & 73.4 & 67.24 \\
Fast-WAM-joint (\textit{Gaussian})
    & $\times$ & 58.22 & 76.71 & \textbf{95.77} & 92.38
    & 64.68 & 74.58 & 81.11 & 77.51 \\
Fast-WAM-joint (\m{}-flow)
    & $\times$ & 70.04 & 78.77 & 85.43 & 91.77
    & 68.12 & 88.38 & 80.79 & 80.58 \\
\bottomrule
\end{tabular*}
\par
\endgroup
\end{table}

\subsection{Generalization under Distribution Shift}

We next evaluate generalization on both WAM and VLA policies using LIBERO-Plus without additional fine-tuning. Table~\ref{tab:main-results} summarizes their performance on LIBERO and LIBERO-Plus. On ImageWAM, \m{}-flow with a 50\% image-latent prior increases LIBERO-Plus success from $82.49\%$ to $88.55\%$, while maintaining similar LIBERO performance ($97.80\%$ versus $98.00\%$). ImageWAM with \m{}-flow achieves state-of-the-art performance on LIBERO-Plus, despite using no additional embodied-data pretraining. Fast-WAM-joint with \m{}-flow reaches $80.58\%$ on LIBERO-Plus, compared with $77.51\%$ for the \textit{Gaussian} sparse-joint reference with the same architecture and one-step sampling budget. On the VLA policy $\pi_{0.5}$, action \m{}-flow increases LIBERO-Plus success from $80.70\%$ to $84.53\%$ and LIBERO success from $97.10\%$ to $99.00\%$.

As shown in the bottom panels of Figure~\ref{fig:training-efficiency_and_generalization}, these generalization gains are already visible early in training. On ImageWAM, \m{}-flow reaches $75.36\%$ LIBERO-Plus success after one epoch, compared with $17.12\%$ for the original model. On Fast-WAM, it reaches $68.55\%$ after two epochs, compared with $47.81\%$ for the \textit{Gaussian} sparse-joint baseline. By epoch four, its success reaches $77.81\%$, already exceeding the baseline's final $77.51\%$.

The perturbation-wise results further show where these improvements are most pronounced. On ImageWAM, the largest gains occur under changes in robot initialization and object layout, where success increases from $52.90\%$ to $73.94\%$ and from $81.31\%$ to $88.39\%$, respectively. Gains also extend to language instructions and visual appearance, including background textures and lighting. Improvements under sensor noise are smaller, while camera-viewpoint performance remains largely unchanged. Taken together, these results show that the benefits of \m{}-flow extend beyond faster learning on LIBERO to stronger generalization under distribution shift across both WAM and VLA policies. We attribute this generalization to two factors: the residual is easier to fit than a full transport, and the executed history now enters as the flow source rather than as a state token concatenated with the visual input, which removes the shortcut that such concatenation often invites.

\begin{figure}[htbp]
\centering
\includegraphics[width=1.0\linewidth]{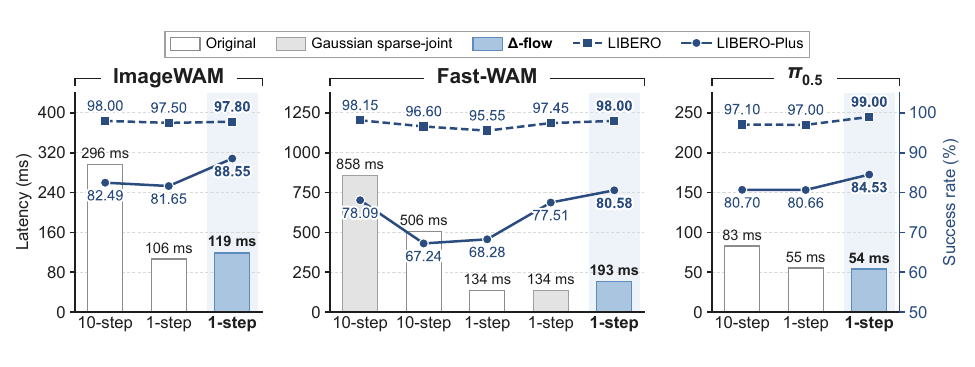}
\caption{\textbf{Inference latency and task success on ImageWAM, Fast-WAM, and $\pi_{0.5}$.}
Bars show mean complete local policy-call latency on an A100 GPU, including online text encoding and visual source construction, with values read from the left axes.
White, gray, and blue bars denote the original models,
Gaussian sparse-joint Fast-WAM, and \m{}-flow, respectively.
Dashed lines with square markers show LIBERO success,
and solid lines with circle markers show LIBERO-Plus success,
with values read from the right axis.}

\label{fig:inference-latency}
\end{figure}
\subsection{Efficient and Smooth Inference}
\label{sec:efficient-smooth-inference}

We measure complete local inference latency on an A100-80GB GPU, with the detailed timing protocol provided in Appendix~\ref{app:evaluation-protocols}. As shown in Figure~\ref{fig:inference-latency}, one-step \m{}-flow reduces inference latency across all three policies. On ImageWAM, mean latency decreases from $296.31$ to $118.87$\,ms relative to ten-step \textit{Gaussian} sampling, giving a $2.49\times$ speedup. The original one-step model takes $106.46$\,ms but does not predict future visual representations. On Fast-WAM, complete one-step generation takes $192.62$\,ms. Compared with ten-step \textit{Gaussian} sparse-joint inference at $857.54$\,ms, this gives a $4.45\times$ speedup. Its one-step \textit{Gaussian} sparse-joint reference takes $133.71$\,ms. On $\pi_{0.5}$, one-step action \m{}-flow takes $53.99$\,ms, compared with $82.91$\,ms for ten-step and $55.34$\,ms for one-step \textit{Gaussian} sampling. This gives a $1.54\times$ speedup over ten-step inference. Thus, \m{}-flow reduces the latency of the complete inference pipeline across WAM and VLA policies through single-step generation.

\begin{figure}[htbp]
\centering
\includegraphics[width=0.9\linewidth]{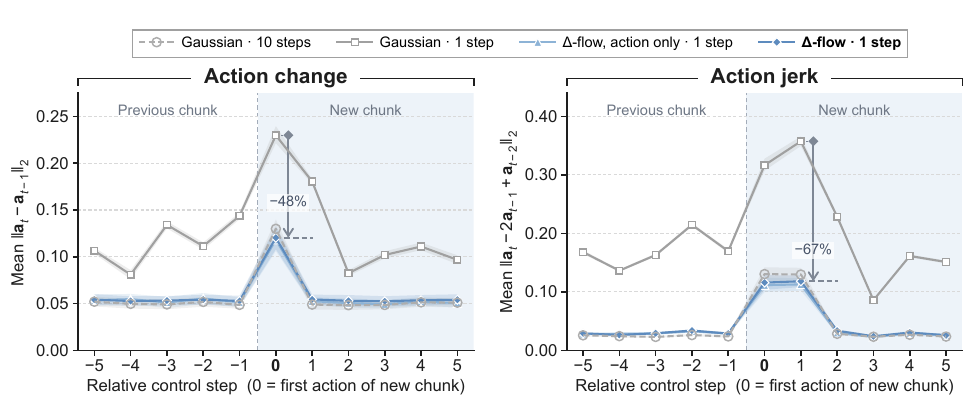}
\caption{\textbf{Action continuity at chunk boundaries on ImageWAM.}
The panels show mean action change (left) and action jerk (right)
around transitions between consecutive action chunks.
The four curves compare \textit{Gaussian} sampling with ten or one step,
one-step action-only \m{}-flow, and one-step complete \m{}-flow.
All four methods are evaluated on the same LIBERO-Long tasks
with identical initial states.
Each policy predicts a 16-action chunk and executes the first 12 actions before replanning.
Step 0 is the first action of the new chunk.
Background shading indicates the new chunk,
and bands around the curves show paired-bootstrap 95\% confidence intervals.
Arrows annotate relative differences in boundary peaks.}

\label{fig:action-continuity}
\end{figure}
We next examine action continuity during replanning. On ImageWAM, we compare ten-step and one-step \textit{Gaussian} sampling, one-step action-only \m{}-flow, and one-step complete \m{}-flow. All methods are evaluated on the same LIBERO-Long tasks with identical initial states. Each policy predicts a 16-action chunk and executes the first 12 actions before replanning. We measure consecutive action change, action jerk computed from second finite differences, and jumps at action-chunk boundaries. Reducing \textit{Gaussian} sampling from ten steps to one increases these measures to $2.19$, $4.60$, and $1.77$ times their ten-step values, respectively. Reducing \textit{Gaussian} sampling steps alone therefore comes at a clear cost to action continuity during replanning.

In contrast, the left and right panels of Figure~\ref{fig:action-continuity} show that complete \m{}-flow reduces boundary peaks in action change and jerk by $48\%$ and $67\%$, respectively, relative to one-step \textit{Gaussian} sampling. Both \m{}-flow variants remain close to the ten-step \textit{Gaussian} reference. Across the matched trajectories, complete \m{}-flow reduces mean action jerk from $0.19399$ to $0.04335$ relative to one-step \textit{Gaussian} sampling, a $4.5\times$ reduction (Table~\ref{tab:trajectory-mean-jerk} in Appendix~\ref{app:evaluation-protocols}). The history-based source provides a temporal reference from recent motion, while endpoint consistency directly supervises the future action chunk predicted in one update. Together, these designs align training with single-step generation, rather than merely reducing the sampling budget at inference. These results suggest that \m{}-flow preserves action continuity by combining a history-based starting point with direct one-step supervision, without relying on repeated sampling updates.

\begin{table*}[htbp]
\centering
\caption{Training-free action-source adaptation on LIBERO and LIBERO-Plus. Only the inference-time source of a frozen policy is changed, without updating model parameters. Success rates are in \%. Steps denotes the number of inference sampling steps.}
\label{tab:training-free-source}

\small
\setlength{\tabcolsep}{5pt}
\renewcommand{\arraystretch}{1.1}

\begin{tabular}{@{}llccc@{}}
\toprule
Policy & Source & Steps & LIBERO & LIBERO-Plus \\
\midrule
ImageWAM & \textit{Gaussian} & 10 & 98.00 & 82.49 \\
ImageWAM & \textit{Gaussian} & 1 & 97.50 & 81.65 \\
ImageWAM & Past action & 1 & \textbf{98.10} & \textbf{82.82} \\
\addlinespace
Fast-WAM & \textit{Gaussian} & 10 & \textbf{96.60} & 67.24 \\
Fast-WAM & \textit{Gaussian} & 1 & 95.55 & \textbf{68.28} \\
Fast-WAM & Past action & 1 & 96.15 & 67.58 \\
\bottomrule
\end{tabular}
\vspace{2pt}
\end{table*}

\subsection{Training-Free Source Adaptation}
\label{sec:training-free-source}

Lastly, we test action-history initialization on the original ImageWAM and Fast-WAM policies trained with Gaussian sources. Both policies are frozen, and only their \textit{Gaussian} action source is replaced at inference. The adaptation is a special case of Eq.~\eqref{eq:difference-source}. We set $\alpha_a=1$ and $\sigma_a=0.5$, with $\mathbf{b}_a$ restricting the noise to the pose dimensions.

As shown in Table~\ref{tab:training-free-source}, replacing the action source at a single sampling step raises ImageWAM's LIBERO success from $97.50\%$ to $98.10\%$. Fast-WAM improves from $95.55\%$ to $96.15\%$ under the same one-step budget. On LIBERO-Plus, ImageWAM improves from $81.65\%$ to $82.82\%$, while Fast-WAM decreases from $68.28\%$ to $67.58\%$. The adaptation is also free at a fixed solver budget, since it adds no network evaluation. Latency follows the complete local policy-call protocol on an A100-80GB GPU, consistent with Section~\ref{sec:efficient-smooth-inference} and Appendix~\ref{app:evaluation-protocols}. These results show that an action-history source transfers to a frozen WAM policy without post-training, supporting one-step inference while largely preserving task success. It further indicates that \m{}-flow can be combined with an already well-trained policy at little cost, yielding immediate inference gains.

\section{Ablation Studies}
\label{sec:ablations}

Building on the performance and efficiency results above, we examine two main design choices in \m{}-flow differently from the conventional flow policy: learning benefits in the action and visual branches and supervising one-step predictions with endpoint consistency. The component and objective ablations use one-step inference. All ablations in this section are conducted on ImageWAM.

\subsection{How Much Does Each Flow Contribute?}
\label{sec:visual-action-contributions}

\begin{table}[htbp]
\centering
\caption{Contributions of the action and visual \m{}-flows on ImageWAM. Change (pp) is the signed LIBERO-Plus difference from the original \textit{Gaussian}-source model.}
\label{tab:flow-components}
\small
\setlength{\tabcolsep}{5pt}
\begin{tabular}{lccc}
\toprule
Configuration & LIBERO & LIBERO-Plus & Change (pp) \\
\midrule
Original ImageWAM & 98.00 & 82.49 & -- \\
\m{}-flow, action only & \textbf{98.40} & 85.82 & $+3.33$ \\
\m{}-flow, action $+$ visual & 97.80 & \textbf{88.55} & $\mathbf{+6.06}$ \\
\bottomrule
\end{tabular}
\end{table}

\textbf{Training efficiency.} We compare the original ImageWAM, action-only \m{}-flow, and complete \m{}-flow to examine how the action and visual components contribute to learning efficiency. As shown in the ImageWAM LIBERO panel of Figure~\ref{fig:training-efficiency_and_generalization}, action-only \m{}-flow first reaches $90\%$ success at epoch four, compared with epoch six for the original model. Applying \m{}-flow to both actions and visual representations achieves over $90\%$ LIBERO success after the first epoch. Thus, action-side \m{}-flow accelerates learning on its own, while extending the formulation to the visual branch further reduces the training required to achieve high task success.\par \textbf{Generalization.} We next compare the selected models on LIBERO-Plus to examine the contributions of the action and visual components under distribution shift. As shown in Table~\ref{tab:flow-components}, action-only \m{}-flow increases LIBERO-Plus success from $82.49\%$ to $85.82\%$. Adding visual \m{}-flow further raises success to $88.55\%$, an additional gain of $2.73$ percentage points. Action-side \m{}-flow already improves learning efficiency and generalization. Applying \m{}-flow to both action and visual generation further accelerates learning and strengthens generalization under distribution shift.

\subsection{Where Does Endpoint Consistency Help?}
\label{sec:endpoint-consistency}

\begin{table}[htbp]
\centering
\caption{Endpoint consistency on ImageWAM. Success rates (\%) are reported for the final trained models. The action-only model provides a reference. The visual endpoint-consistency ablation uses the 80\% image-latent prior.}
\label{tab:iw-endpoint}
\small
\setlength{\tabcolsep}{5pt}
\begin{tabular}{lcc}
\toprule
Configuration & LIBERO & LIBERO-Plus \\
\midrule
\m{}-flow, action only & 98.10 & 85.75 \\
\addlinespace[2pt]
\m{}-flow, action $+$ visual (80\% image-latent prior) & 97.70 & \textbf{87.37} \\
\quad w/o $\mathcal{L}_{\mathrm{EC}}^v$ & \textbf{98.60} & 84.96 \\
\bottomrule
\end{tabular}
\end{table}

Flow matching (Eq.~\eqref{eq:difference-fm}) supervises velocity along the interpolation path, whereas endpoint consistency (Eq.~\eqref{eq:endpoint-consistency}) directly constrains the prediction obtained after one update from the source. We remove the visual endpoint loss to assess its contribution. Table~\ref{tab:iw-endpoint} summarizes the experimental results. 

Removing the visual endpoint loss $\mathcal{L}_{\mathrm{EC}}^v$ from the $80\%$ image-latent prior configuration lowers LIBERO-Plus success from $87.37\%$ to $84.96\%$.
The corresponding change on standard LIBERO is $+0.90$ percentage points. The visual comparison shows that endpoint consistency improves success under distribution shift, even though standard LIBERO success changes little.

\section{Empirical Studies on Policy Design}
\label{sec:discussion}

Beyond the components of \m{}-flow itself, four further design choices must be settled before a history-based formulation can be deployed, and together they define the configuration used in our final models: \textbf{(1)} how much future visual prediction control actually needs; \textbf{(2)} whether gradients should be isolated between the world model and the action head; \textbf{(3)} whether the action branch still requires flow-matching supervision once its source is past-informed; and \textbf{(4)} how the action representation affects generalization. Although we settle them here for \m{}-flow, the same questions arise in any policy that couples world modeling with action generation, and we expect the comparisons below to be informative beyond this work.

\begin{figure}[htbp]
\centering
\includegraphics[width=0.9\linewidth]{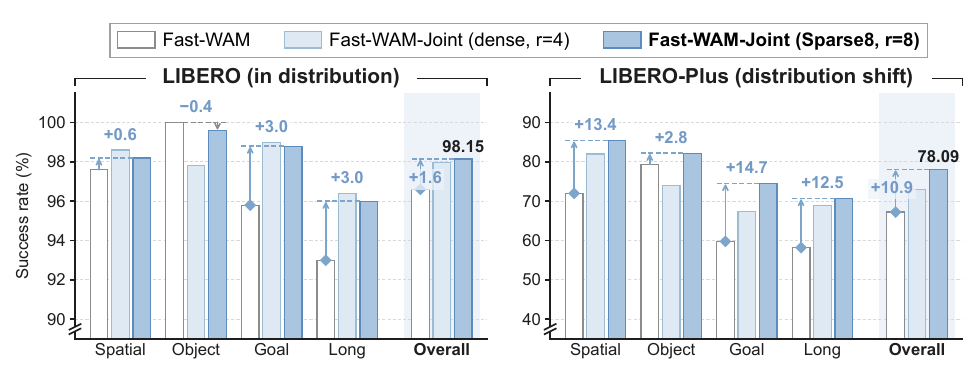}
\caption{Success rates (\%) of Fast-WAM variants on LIBERO and LIBERO-Plus.
All models are jointly trained on the four LIBERO suites.}
\label{fig:fastwam-variants}
\end{figure}
\subsection{How Much Future Does Control Need?}
\label{sec:discussion-information-coupling}

\textbf{Future visual context.} Fast-WAM omits future video generation at inference, whereas Fast-WAM-Joint jointly predicts future visual representations and actions. As shown in Figure~\ref{fig:fastwam-variants}, the dense joint variant improves success from $96.60\%$ to $97.95\%$ on LIBERO and from $67.24\%$ to $72.90\%$ on LIBERO-Plus. 
The $5.66$-percentage-point increase on LIBERO-Plus, compared with $1.35$ points on LIBERO, shows that future visual prediction benefits control particularly under distribution shift.

\textbf{Sparse future prediction.}
 Over the same 32-step horizon, Fast-WAM-Joint-Sparse (Sparse8, $r=8$) predicts four future frames instead of the eight used by the dense joint variant ($r=4$), yet increases success from $97.95\%$ to $98.15\%$ on LIBERO and from $72.90\%$ to $78.09\%$ on LIBERO-Plus (Figure~\ref{fig:fastwam-variants}). 
These results favor sparse temporal coverage over dense frame prediction.
Additional intermediate frames can introduce redundancy without providing more useful information for control. 
Our ImageWAM-based model predicts only a single future image and also achieves strong task performance (Table~\ref{tab:main-results}). Moreover, sparse future prediction can further reduce inference latency.

\begin{figure}[htbp]
\centering
\includegraphics[width=0.8\linewidth]{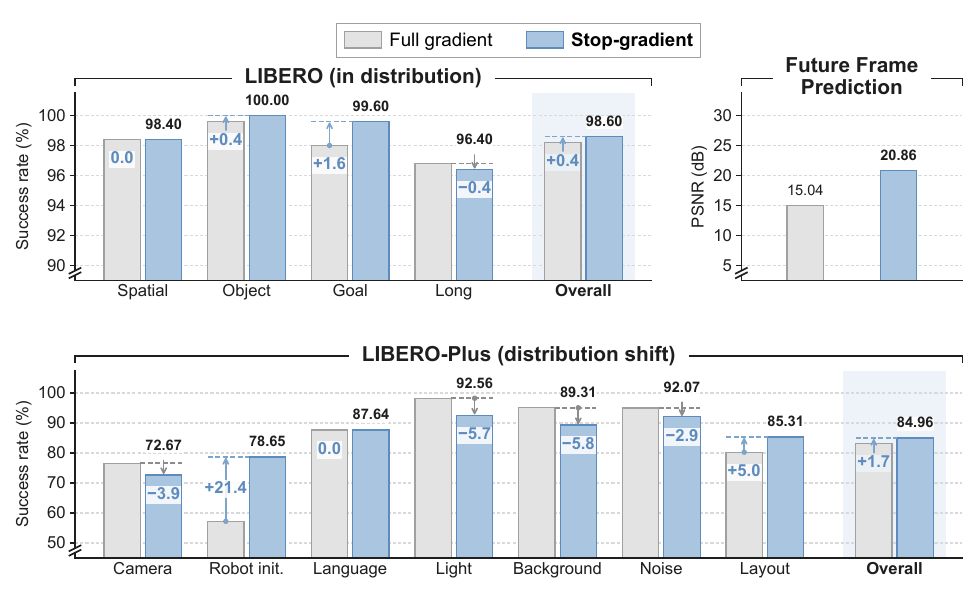}
\caption{\textbf{Action-gradient transfer into the visual branch.}
Comparison across LIBERO suites, future-image prediction PSNR, and LIBERO-Plus perturbations.
Black numbers in the success-rate panels report stop-gradient success rates.
Blue annotations show differences from full gradient in percentage points.}
\label{fig:iw_visual_stopgrad}
\end{figure}
\subsection{Should the Action Loss Gradient Reach the World Model?}

Having examined how future visual prediction supports action generation, we next investigate whether action-loss gradients should update the visual branch. We compare full gradient with stop-gradient without visual endpoint consistency in our ImageWAM-based model. Full gradient allows action-loss gradients through the visual-to-action connection, whereas stop-gradient blocks them while preserving visual conditioning. Future-image prediction is evaluated on five tasks each from LIBERO-Goal and LIBERO-Object, using 250 rollouts per configuration. One-step predictions are compared with ground-truth images 16 control steps ahead.

As shown in Figure~\ref{fig:iw_visual_stopgrad}, stop-gradient achieves higher future-image PSNR and higher overall LIBERO-Plus success. With stop-gradient, the WAM visual branch learns future visual changes more accurately under visual supervision. Action generation conditioned on its intermediate representations also achieves higher overall success. These results suggest that allowing the visual branch to learn independently of action-loss gradients can benefit both future visual prediction and action generation in WAMs.

\begin{figure}[t]
\centering
\includegraphics[width=0.9\linewidth]{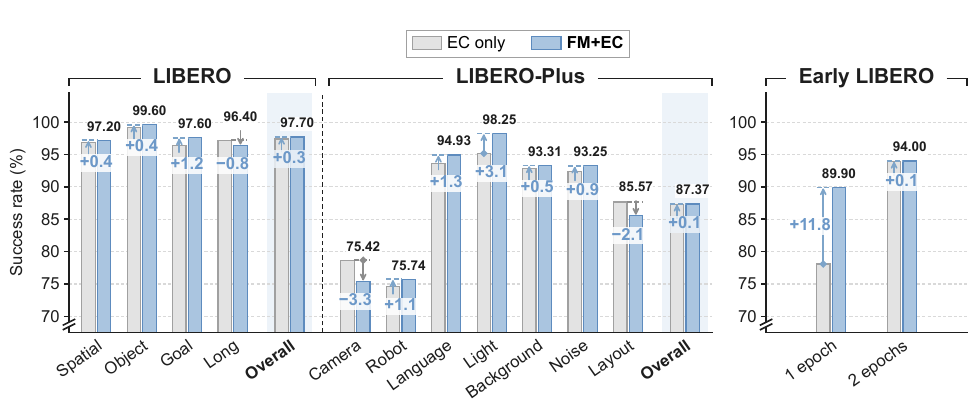}

\caption{Action objectives in ImageWAM (success rate, \%). Both models use the 80\% image-latent prior. The action FM/EC weights are $1.0/0.5$ for FM+EC and $0/1.0$ for EC only; the visual objectives are unchanged. EC is L1 endpoint consistency.}
\label{fig:iw-consistency-only}
\end{figure}
\subsection{Is Action Flow Matching Still Necessary?}
\label{sec:discussion-consistency-only}

We next examine whether endpoint consistency alone can supervise the action branch in ImageWAM with the $80\%$ image-latent prior. We remove action flow matching and increase the action endpoint-consistency weight from $0.5$ to $1.0$, while keeping the visual objectives unchanged. Detailed training settings are provided in Appendix~\ref{app:action-objective}. As shown in the left and middle panels of Figure~\ref{fig:iw-consistency-only}, endpoint-only training closely matches the combined objective. It achieves $97.40\%$ success on LIBERO and $87.30\%$ on LIBERO-Plus, compared with $97.70\%$ and $87.37\%$, respectively. Endpoint consistency alone thus supports accurate one-step action prediction without supervising velocities along the interpolation path.

Flow matching nevertheless improves early learning, as shown in the right panel of Figure~\ref{fig:iw-consistency-only}. After one epoch, the combined objective achieves $89.90\%$ LIBERO success, compared with $78.10\%$ for endpoint consistency alone. By the second epoch, the two configurations achieve nearly identical success rates. Flow matching therefore helps the action branch learn faster early on, while endpoint consistency alone nearly matches its final one-step performance. We retain both objectives to combine faster early learning with accurate one-step prediction.

\subsection{How to Choose Action Representation?}
\label{sec:discussion-action-space}

\begin{figure}[htbp]
\centering
\includegraphics[width=0.9\linewidth]{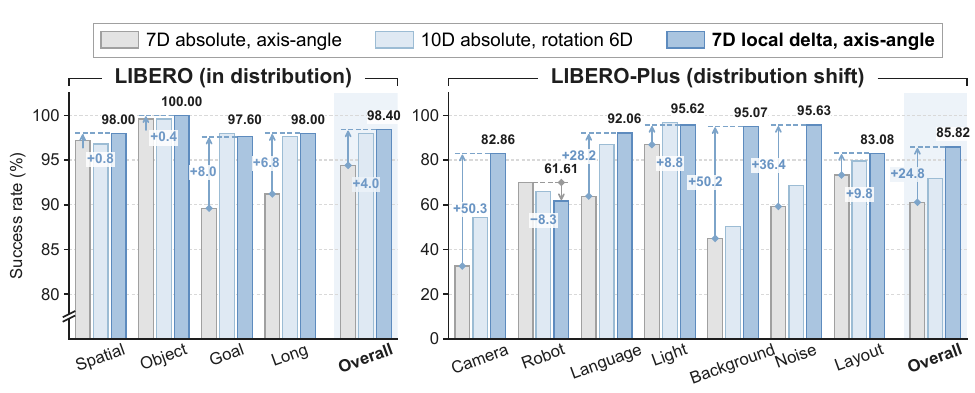}
\caption{\textbf{Action representations on LIBERO and LIBERO-Plus.}
We compare 7D local delta actions with axis-angle rotations,
7D absolute actions with axis-angle rotations,
and 10D absolute actions with 6D rotations.
The left panel shows success rates across LIBERO suites,
and the right panel shows success rates across LIBERO-Plus perturbation categories.
Black numbers report local-delta success. Blue annotations show local delta minus 7D absolute in percentage points.}
\label{fig:iw-action-space}
\end{figure}

The action representation defines the space in which \m{}-flow learns history-to-future changes. Figure~\ref{fig:iw-action-space} compares three representations: 7D local delta with axis-angle rotations, 7D absolute with axis-angle rotations, and 10D absolute with rotation-6D. The three control spaces are defined in Appendix~\ref{app:action-representation}.

With axis-angle rotations, the local-delta representation achieves $98.40\%$ success on LIBERO and $85.82\%$ on LIBERO-Plus, compared with $94.40\%$ and $61.07\%$ for the absolute representation. The difference is particularly pronounced under camera-viewpoint, background-texture, and sensor-noise perturbations. Absolute actions, however, perform better under robot-initial-state perturbations, indicating that the relative performance also depends on the type of distribution shift.

Within the absolute-action configurations, rotation-6D increases success from $94.40\%$ to $98.00\%$ on LIBERO and from $61.07\%$ to $71.67\%$ on LIBERO-Plus. It performs better in six of the seven perturbation categories, with robot initial states as the exception. Although the 10D absolute representation approaches local-delta performance on standard LIBERO, its LIBERO-Plus success remains lower, at $71.67\%$ versus $85.82\%$. %
Overall, local-delta actions achieve the highest aggregate success on both benchmarks, while rotation-6D performs better than axis-angle within the absolute-action configurations. These results show that action representation deserves consideration alongside source construction when learning history-to-future changes.

\FloatBarrier
\section{Conclusion}
\label{sec:conclusion}

We revisited a design choice that flow-based robot policies take for granted: 
where generation begins. Visual states and actions both
change continuously, so the recent past is already close to the future a policy
must produce. Taking it as the flow source leaves only a small residual \m{} to
learn. \m{}-flow builds such a source by perturbing the history with mild noise.
The change touches mainly the source, so it integrates into existing backbones without new modules or extra data. ImageWAM reaches $90\%$ LIBERO success in one epoch rather than six, while Fast-WAM reaches $95\%$ in two epochs rather than eight. Complete one-step inference gives $2.49\times$ and $4.45\times$ speedups over their ten-step Gaussian references. LIBERO-Plus success improves by $6.06$ points on ImageWAM and $3.07$ points over the Gaussian sparse-joint Fast-WAM reference. On $\pi_{0.5}$, the construction applies to actions alone. With LoRA fine-tuning, its first-epoch result exceeds the baseline's five-epoch result, and it reaches $98.00\%$ on LIBERO after five epochs. The approach also preserves action continuity at one step, and action-source replacement can improve a frozen Gaussian-trained WAM policy without retraining. These results support history-based sources as a practical design choice across robot foundation backbones.

\section{Future Work}
\label{sec:future-work}

The noise scale $\sigma_m$ is a hyperparameter, although in practice it requires
no tuning: all three backbones use the same action-source noise scale. %
Still, the history is not equally reliable throughout a trajectory~\citep{jia2026marspolicy}. Adapting $\sigma_m$ per sample or per
timestep, from the observation, from how fast the scene is changing, or learned
together with the vector field, is a promising extension.
Validating \m{}-flow on more VLA backbones and on physical robots also remains
future work.


\bibliographystyle{assets/plainnat}
\bibliography{paper}

\begin{thebibliography}{37}
\providecommand{\natexlab}[1]{#1}
\providecommand{\url}[1]{\texttt{#1}}
\expandafter\ifx\csname urlstyle\endcsname\relax
  \providecommand{\doi}[1]{doi: #1}\else
  \providecommand{\doi}{doi: \begingroup \urlstyle{rm}\Url}\fi

\bibitem[An et~al.(2026)An, Jia, Li, Li, Zhou, Liu, Lyu, Bai, Guo, Li, and Yang]{an2026feedback}
Tuo An, Jindou Jia, Gen Li, et~al.
\newblock Feedback world model enables precise guidance of diffusion policy.
\newblock In \emph{Advances in Neural Information Processing Systems}, Sydney, Australia, 2026.

\bibitem[Bai et~al.(2026)Bai, Jia, Hu, Li, Chen, An, Zuo, and Yang]{bai2026flash}
Jiaqi Bai, Jindou Jia, Yuxuan Hu, et~al.
\newblock Flash: Efficient visuomotor policy via sparse sampling.
\newblock In \emph{Advances in Neural Information Processing Systems}, Sydney, Australia, 2026.

\bibitem[Bi et~al.(2025)Bi, Tan, Xie, Wang, Huang, Liu, Zhao, Feng, Xiang, Rong, Zhao, Liu, Su, Ma, Su, and Zhu]{bi2025motus}
Hongzhe Bi, Hengkai Tan, Shenghao Xie, et~al.
\newblock Motus: A unified latent action world model.
\newblock \emph{arXiv preprint arXiv:2512.13030}, 2025.
\newblock \url{https://arxiv.org/abs/2512.13030}.

\bibitem[Black et~al.(2025{\natexlab{a}})Black, Brown, Darpinian, Dhabalia, Driess, Esmail, Equi, Finn, Fusai, Galliker, Ghosh, Groom, Hausman, Ichter, Jakubczak, Jones, Ke, LeBlanc, Levine, Li-Bell, Mothukuri, Nair, Pertsch, Ren, Shi, Smith, Springenberg, Stachowicz, Tanner, Vuong, Walke, Walling, Wang, Yu, and Zhilinsky]{intelligence2025pi05}
Kevin Black, Noah Brown, James Darpinian, et~al.
\newblock {$\pi_{0.5}$}: a vision-language-action model with open-world generalization.
\newblock In \emph{Proceedings of The 9th Conference on Robot Learning}, volume 305 of \emph{Proceedings of Machine Learning Research}, pages 17--40. PMLR, 2025{\natexlab{a}}.

\bibitem[Black et~al.(2025{\natexlab{b}})Black, Brown, Driess, Esmail, Equi, Finn, Fusai, Groom, Hausman, Ichter, Jakubczak, Jones, Ke, Levine, Li-Bell, Mothukuri, Nair, Pertsch, Shi, Smith, Tanner, Vuong, Walling, Wang, and Zhilinsky]{black2024pi0}
Kevin Black, Noah Brown, Danny Driess, et~al.
\newblock {$\pi_0$}: A vision-language-action flow model for general robot control.
\newblock In \emph{Proceedings of Robotics: Science and Systems}, 2025{\natexlab{b}}.

\bibitem[Bu et~al.(2025)Bu, Yang, Cai, Gao, Ren, Yao, Luo, and Li]{bu2025univla}
Qingwen Bu, Yanting Yang, Jisong Cai, et~al.
\newblock {UniVLA}: Learning to act anywhere with task-centric latent actions.
\newblock In \emph{Proceedings of Robotics: Science and Systems}, 2025.
\newblock \doi{10.48550/arXiv.2505.06111}.
\newblock \url{https://arxiv.org/abs/2505.06111}.

\bibitem[Chen et~al.(2024)Chen, Lim, Lin, Chen, and Soh]{chen2024bridger}
Kaiqi Chen, Eugene Lim, Kelvin Lin, et~al.
\newblock Don't start from scratch: Behavioral refinement via interpolant-based policy diffusion.
\newblock In \emph{Proceedings of Robotics: Science and Systems}, Delft, Netherlands, July 2024.

\bibitem[Chen et~al.(2026)Chen, Yang, Tang, Huo, Lin, Wu, Liu, Chen, Zheng, Yuan, Li, Wang, Qi, Hu, Mei, Xuan, Yang, Zhu, Xu, Ma, and Chang]{chen2026abotm05}
Ronghan Chen, Yandan Yang, Zuojin Tang, et~al.
\newblock {ABot-M0.5}: Unified mobility-and-manipulation world action model.
\newblock \emph{arXiv preprint arXiv:2607.00678}, 2026.
\newblock \url{https://arxiv.org/abs/2607.00678}.

\bibitem[Davtyan et~al.(2023)Davtyan, Sameni, and Favaro]{Davtyan_2023_ICCV}
Aram Davtyan, Sepehr Sameni, and Paolo Favaro.
\newblock Efficient video prediction via sparsely conditioned flow matching.
\newblock In \emph{Proceedings of the IEEE/CVF International Conference on Computer Vision}, pages 23263--23274, October 2023.

\bibitem[Dong et~al.(2026)Dong, Lyu, Liu, Wang, Liang, Zhang, Tu, Li, Zhao, Ding, Zhang, Han, Sebe, Khan, Khan, Shah, Torr, Yang, and Tao]{WM_Survey}
Jiahua Dong, Qi~Lyu, Baichen Liu, et~al.
\newblock Learning to model the world: A survey of world models in artificial intelligence.
\newblock \emph{TechRxiv}, 2026.

\bibitem[Fei et~al.(2026)Fei, Wang, Shi, Dai, Cai, Qian, Ji, He, Zhang, Fei, Fu, Gong, and Qiu]{fei2026liberoplus}
Senyu Fei, Siyin Wang, Junhao Shi, et~al.
\newblock {LIBERO-Plus}: A progressive robustness benchmark for visual-language-action models.
\newblock In \emph{Proceedings of the IEEE/CVF Conference on Computer Vision and Pattern Recognition (CVPR)}, pages 38574--38583, 2026.

\bibitem[Hou et~al.(2026)Hou, Li, Jia, An, Guo, Leng, Geng, Ze, Harada, Torr, et~al.]{hou2026world}
Bohan Hou, Gen Li, Jindou Jia, et~al.
\newblock World model for robot learning: A comprehensive survey.
\newblock \emph{The International Journal of Robotics Research}, 2026.

\bibitem[Hu et~al.(2022)Hu, Shen, Wallis, Allen-Zhu, Li, Wang, Wang, Chen, et~al.]{hu2022lora}
Edward~J Hu, Yelong Shen, Phillip Wallis, et~al.
\newblock {LoRA}: Low-rank adaptation of large language models.
\newblock In \emph{Proceedings of International Conference on Learning Representations}, 2022.

\bibitem[Jia et~al.(2026{\natexlab{a}})Jia, An, Hu, Li, Li, Hou, Chen, Bai, Lyu, and Yang]{jia2026marspolicy}
Jindou Jia, Tuo An, Yuxuan Hu, et~al.
\newblock {MARS} policy: multimodality only when it matters.
\newblock \emph{arXiv preprint arXiv:2605.29766}, 2026{\natexlab{a}}.

\bibitem[Jia et~al.(2026{\natexlab{b}})Jia, Li, Chen, An, Hu, Li, Guo, and Yang]{a2a2026}
Jindou Jia, Gen Li, Xiangyu Chen, et~al.
\newblock {Action-to-Action Flow Matching}.
\newblock In \emph{Proceedings of Robotics: Science and Systems}, July 2026{\natexlab{b}}.

\bibitem[Kim et~al.(2025)Kim, Pertsch, Karamcheti, Xiao, Balakrishna, Nair, Rafailov, Foster, Sanketi, Vuong, Kollar, Burchfiel, Tedrake, Sadigh, Levine, Liang, and Finn]{kim2025openvla}
Moo~Jin Kim, Karl Pertsch, Siddharth Karamcheti, et~al.
\newblock {OpenVLA}: An open-source vision-language-action model.
\newblock In \emph{Proceedings of The 8th Conference on Robot Learning}, volume 270 of \emph{Proceedings of Machine Learning Research}, pages 2679--2713. PMLR, 2025.

\bibitem[Kim et~al.(2026)Kim, Gao, Lin, Lin, Ge, Lam, Liang, Song, Liu, Finn, and Gu]{kim2026cosmospolicy}
Moo~Jin Kim, Yihuai Gao, Tsung-Yi Lin, et~al.
\newblock Cosmos policy: Fine-tuning video models for visuomotor control and planning.
\newblock \emph{arXiv preprint arXiv:2601.16163}, 2026.
\newblock \url{https://arxiv.org/abs/2601.16163}.

\bibitem[Li et~al.(2026{\natexlab{a}})Li, Cong, Wang, Xia, Huang, Zhang, Xu, and Dai]{li2026step}
Jinhao Li, Yuxuan Cong, Yingqiao Wang, et~al.
\newblock {STEP}: Warm-started visuomotor policies with spatiotemporal consistency prediction.
\newblock In \emph{Proceedings of the 43rd International Conference on Machine Learning}, volume 306 of \emph{Proceedings of Machine Learning Research}, pages 66501--66525. PMLR, 2026{\natexlab{a}}.

\bibitem[Li et~al.(2026{\natexlab{b}})Li, Zhang, Luo, Yang, Wang, Han, Yu, Gao, Xue, Zhu, Shen, and Xu]{lingbotva}
Lin Li, Qihang Zhang, Yiming Luo, et~al.
\newblock Causal world modeling for robot control.
\newblock \emph{CoRR}, abs/2601.21998, January 2026{\natexlab{b}}.

\bibitem[Lipman et~al.(2023)Lipman, Chen, Ben-Hamu, Nickel, and Le]{lipman2023flow}
Yaron Lipman, Ricky T.~Q. Chen, Heli Ben-Hamu, et~al.
\newblock Flow matching for generative modeling.
\newblock In \emph{The Eleventh International Conference on Learning Representations}, 2023.

\bibitem[Liu et~al.(2023{\natexlab{a}})Liu, Zhu, Gao, Feng, Liu, Zhu, and Stone]{liu2023libero}
Bo~Liu, Yifeng Zhu, Chongkai Gao, et~al.
\newblock {LIBERO}: Benchmarking knowledge transfer for lifelong robot learning.
\newblock In \emph{Proceedings of Advances in Neural Information Processing Systems}, volume~36, pages 44776--44791, 2023{\natexlab{a}}.

\bibitem[Liu and Ritschel(2025)]{Liu_2025_ICCV}
Chen Liu and Tobias Ritschel.
\newblock Generative video bi-flow.
\newblock In \emph{Proceedings of the IEEE/CVF International Conference on Computer Vision}, pages 19363--19372, October 2025.

\bibitem[Liu et~al.(2023{\natexlab{b}})Liu, Vahdat, Huang, Theodorou, Nie, and Anandkumar]{pmlr-v202-liu23ai}
Guan-Horng Liu, Arash Vahdat, De-An Huang, et~al.
\newblock {{I}$^2${SB}: Image-to-Image Schr{\"o}dinger Bridge}.
\newblock In \emph{Proceedings of the 40th International Conference on Machine Learning}, volume 202 of \emph{Proceedings of Machine Learning Research}, pages 22042--22062. PMLR, 2023{\natexlab{b}}.

\bibitem[Lu et~al.(2022)Lu, Zhou, Bao, Chen, Li, and Zhu]{lu2022dpmsolver}
Cheng Lu, Yuhao Zhou, Fan Bao, et~al.
\newblock {DPM-Solver}: A fast {ODE} solver for diffusion probabilistic model sampling in around 10 steps.
\newblock In \emph{Advances in Neural Information Processing Systems}, volume~35, 2022.

\bibitem[Pertsch et~al.(2025)Pertsch, Stachowicz, Ichter, Driess, Nair, Vuong, Mees, Finn, and Levine]{pertsch2025fast}
Karl Pertsch, Kyle Stachowicz, Brian Ichter, et~al.
\newblock {FAST}: Efficient action tokenization for vision-language-action models.
\newblock In \emph{Proceedings of Robotics: Science and Systems}, 2025.

\bibitem[Prasad et~al.(2024)Prasad, Lin, Wu, Zhou, and Bohg]{prasad2024consistencypolicy}
Aaditya Prasad, Kevin Lin, Jimmy Wu, et~al.
\newblock Consistency policy: Accelerated visuomotor policies via consistency distillation.
\newblock In \emph{Proceedings of Robotics: Science and Systems}, 2024.

\bibitem[Reuss et~al.(2025)Reuss, Zhou, R\"{u}hle, Ya\u{g}murlu, Otto, and Lioutikov]{reuss2025flower}
Moritz Reuss, Hongyi Zhou, Marcel R\"{u}hle, et~al.
\newblock {FLOWER}: Democratizing generalist robot policies with efficient vision-language-flow models.
\newblock In \emph{Proceedings of The 9th Conference on Robot Learning}, volume 305 of \emph{Proceedings of Machine Learning Research}, pages 3736--3761. PMLR, 2025.

\bibitem[Scholz and Turner(2025)]{warmstarts}
Jonas Scholz and Richard~E Turner.
\newblock Warm starts accelerate conditional diffusion.
\newblock \emph{arXiv preprint arXiv:2507.09212}, 2025.

\bibitem[Song et~al.(2023)Song, Dhariwal, Chen, and Sutskever]{song2023consistency}
Yang Song, Prafulla Dhariwal, Mark Chen, et~al.
\newblock Consistency models.
\newblock In \emph{Proceedings of the 40th International Conference on Machine Learning}, volume 202 of \emph{Proceedings of Machine Learning Research}, pages 32211--32252. PMLR, 2023.

\bibitem[Yang et~al.(2026)Yang, Zhang, Yuan, Wan, Zhang, Hu, and Yang]{yang2026tri}
Jinghan Yang, Yunchao Zhang, Wang Yuan, et~al.
\newblock Tri-info: Generalizable, interpretable failure prediction for vla models via information theory.
\newblock \emph{arXiv preprint arXiv:2606.19998}, 2026.

\bibitem[Ye et~al.(2026)Ye, Ge, Zheng, Gao, Yu, Kurian, Indupuru, Tan, Zhu, Xiang, Malik, Lee, Liang, Ranawaka, Gu, Xu, Wang, Hu, Narayan, Bjorck, Wang, Kim, Niu, Zheng, Xie, Wu, Wang, Julian, Xu, Du, Chebotar, Reed, Kautz, Zhu, Fan, and Jang]{dreamzero}
Seonghyeon Ye, Yunhao Ge, Kaiyuan Zheng, et~al.
\newblock World action models are zero-shot policies.
\newblock \emph{CoRR}, abs/2602.15922, February 2026.

\bibitem[Yuan et~al.(2026)Yuan, Dong, Liu, and Zhao]{fastwam}
Tianyuan Yuan, Zibin Dong, Yicheng Liu, et~al.
\newblock {Fast-WAM}: Do world action models need test-time future imagination?
\newblock In \emph{Advances in Neural Information Processing Systems}, 2026.

\bibitem[Zhang et~al.(2025)Zhang, Liu, Fan, Liu, Zeng, and Liu]{zhang2024flowpolicy}
Qinglun Zhang, Zhen Liu, Haoqiang Fan, et~al.
\newblock {FlowPolicy}: Enabling fast and robust 3d flow-based policy via consistency flow matching for robot manipulation.
\newblock \emph{Proceedings of the AAAI Conference on Artificial Intelligence}, 39\penalty0 (14):\penalty0 14754--14762, 2025.

\bibitem[Zhang et~al.(2026{\natexlab{a}})Zhang, Zhang, Qi, Zhang, Lin, Zhang, Mu, Yang, Zeng, and Jin]{imagewam}
Yuyang Zhang, Wenyao Zhang, Zekun Qi, et~al.
\newblock {ImageWAM}: Do world action models really need video generation, or just image editing?
\newblock In \emph{Proceedings of the 10th Conference on Robot Learning}, 2026{\natexlab{a}}.

\bibitem[Zhang et~al.(2026{\natexlab{b}})Zhang, Li, Rahmati, Yang, Ma, Rasouli, Pakdamansavoji, Wu, Zhang, Cao, Wen, Wang, Quan, and Zhang]{zhang2026wamrobustness}
Zhanguang Zhang, Zhiyuan Li, Behnam Rahmati, et~al.
\newblock Do world action models generalize better than {VLAs}? a robustness study.
\newblock \emph{arXiv preprint arXiv:2603.22078}, 2026{\natexlab{b}}.
\newblock \url{https://arxiv.org/abs/2603.22078}.

\bibitem[Zhou et~al.(2019)Zhou, Barnes, Lu, Yang, and Li]{zhou2019rotation6d}
Yi~Zhou, Connelly Barnes, Jingwan Lu, et~al.
\newblock On the continuity of rotation representations in neural networks.
\newblock In \emph{Proceedings of the IEEE/CVF Conference on Computer Vision and Pattern Recognition}, pages 5745--5753, 2019.

\bibitem[Zitkovich et~al.(2023)Zitkovich, Yu, Xu, Xu, Xiao, Xia, Wu, Wohlhart, Welker, Wahid, Vuong, Vanhoucke, Tran, Soricut, Singh, Singh, Sermanet, Sanketi, Salazar, Ryoo, Reymann, Rao, Pertsch, Mordatch, Michalewski, Lu, Levine, Lee, Lee, Leal, Kuang, Kalashnikov, Julian, Joshi, Irpan, Ichter, Hsu, Herzog, Hausman, Gopalakrishnan, Fu, Florence, Finn, Dubey, Driess, Ding, Choromanski, Chen, Chebotar, Carbajal, Brown, Brohan, Arenas, and Han]{zitkovich2023rt2}
Brianna Zitkovich, Tianhe Yu, Sichun Xu, et~al.
\newblock {RT-2}: Vision-language-action models transfer web knowledge to robotic control.
\newblock In \emph{Proceedings of The 7th Conference on Robot Learning}, volume 229 of \emph{Proceedings of Machine Learning Research}, pages 2165--2183. PMLR, 2023.

\end{thebibliography}


\clearpage \appendix
\raggedbottom

\section{Training and Implementation Details}
\label{app:training-implementation}

Table~\ref{tab:training-hyperparameters} summarizes the training hyperparameters. Model-specific inputs, sources, and objectives are described below.

\begin{table}[H]
\centering
\small
\setlength{\tabcolsep}{9pt}
\caption{Training hyperparameters for ImageWAM, Fast-WAM-joint, and $\pi_{0.5}$. Full and LoRA denote full-parameter and LoRA fine-tuning, respectively. Batch size is global.}
\label{tab:training-hyperparameters}
\begin{tabular}{lcccc}
\toprule
Setting & ImageWAM & Fast-WAM-joint & $\pi_{0.5}$ (Full) & $\pi_{0.5}$ (LoRA) \\
\midrule
Training hardware & $8\times$ A100 & $8\times$ A100 & $8\times$ A100 & $8\times$ A100 \\
Batch size & 80 & 128 & 384 & 384 \\
Training epochs & 10 & 10 & 10 & 5 \\
Optimizer & AdamW & AdamW & AdamW & AdamW \\
Peak learning rate & $10^{-4}$ & $10^{-4}$ & $5\times10^{-5}$ & $5\times10^{-5}$ \\
Warmup & 5\% of steps & 5\% of steps & 100 steps & 100 steps \\
Learning-rate decay & Cosine & Cosine & Cosine & Cosine \\
Final learning rate & $10^{-6}$ & $10^{-6}$ & $5\times10^{-6}$ & $5\times10^{-6}$ \\
\bottomrule
\end{tabular}
\end{table}

ImageWAM and Fast-WAM-joint use AdamW with $(\beta_1,\beta_2)=(0.9,0.95)$ and weight decay $0.01$. Both use one gradient-accumulation step; their per-device batch sizes are 10 and 16, respectively. Full-parameter and LoRA fine-tuning of $\pi_{0.5}$ use JAX FSDP without exponential moving average.

\subsection{ImageWAM Setup}
\label{app:imagewam-setup}

\paragraph{Backbone, data, and inputs.}
We implement ImageWAM in PyTorch with the FLUX.2-klein-base-4B image-editing backbone and an action DiT~\citep{imagewam}. We update both DiTs and the proprioception encoder while freezing the VAE and text encoder. The action DiT starts from the ImageWAM action-backbone initialization checkpoint, with its initial output-head weights multiplied by $0.1$. Training uses the four LIBERO suites in no-no-op LeRobot format. External-camera and wrist-camera images are resized to $224\times224$ and concatenated horizontally into a $224\times448$ input. Each example pairs a language instruction and the current image at time $t$ with the future image at $t+16$ and a 16-action target chunk. Future images are training targets, not observations supplied at deployment. The action expert uses layer-wise visual key--value features, detached at the visual-to-action connection, as illustrated in Figure~\ref{fig:delta0_framework} (left). Table~\ref{tab:imagewam-temporal-layout} summarizes the temporal layout and inference settings.

\begin{table}[H]
\centering
\small
\caption{Temporal layout and inference settings of ImageWAM with \m{}-flow. Future-frame offsets are measured in environment steps relative to decision time $t$.}
\label{tab:imagewam-temporal-layout}
\begin{tabular}{lc}
\toprule
Setting & ImageWAM ($\mathbf{\Delta}$-flow) \\
\midrule
Prediction horizon & 16 steps \\
Action chunk length & 16 commands \\
Action history length & 16 commands \\
Executed actions before replanning & 12 commands \\
Future frame offset & $16$ \\
Number of future frames & 1 \\
Visual sampling steps & 1 \\
Action sampling steps & 1 \\
\bottomrule
\end{tabular}
\end{table}

\paragraph{Sources and objectives.}
The action history contains the previous 16 executed commands, normalized using the run's min/max statistics. We use $\alpha_a=1$ and $\sigma_a=0.5$ in Eq.~\eqref{eq:difference-source}; $\mathbf b_a$ excludes the gripper from source noise. The principal visual source is $(\mathbf h_v+\bm\epsilon_v)/\sqrt{2}$, a 50\% image-latent prior. The percentage is the relative squared coefficient $\alpha_v^2/(\alpha_v^2+\sigma_v^2)$, not a fraction of latent tokens. The 80\% variant uses $\mathbf h_v+0.5\bm\epsilon_v$. With $\lambda_a=1$, $\lambda_v=0.5$, and $\lambda_{\mathrm{EC}}^a=\lambda_{\mathrm{EC}}^v=0.5$, Eq.~\eqref{eq:delta-zero-objective} becomes
\begin{equation}
\mathcal L_{\mathrm{WAM}}=\mathcal L_{\mathrm{FM}}^a+0.5\mathcal L_{\mathrm{EC}}^a+0.5\mathcal L_{\mathrm{FM}}^v+0.25\mathcal L_{\mathrm{EC}}^v.
\end{equation}
Removing an endpoint-consistency loss sets its coefficient to zero. In the endpoint-only action comparison, action FM and EC weights change from $(1,0.5)$ to $(0,1)$; both runs use the 80\% image-latent prior and the same visual objectives.

\paragraph{Inference and evaluation.}
The complete model uses one visual and one action Euler update. It predicts a 16-action chunk and executes 12 actions before replanning from a new observation and updated history. 
 Aggregation and timing procedures are specified in Appendix~\ref{app:evaluation-protocols}; frozen-policy source replacement is described in Appendix~\ref{app:training-free}.

\subsection{Fast-WAM Setup}
\label{app:fastwam-setup}

\paragraph{Backbone, data, and inputs.}
We implement the joint-attention variant of Fast-WAM in PyTorch with the Wan2.2-TI2V-5B video backbone and an action DiT~\citep{fastwam}. Training uses the four LIBERO suites in no-no-op LeRobot format. External-camera and wrist-camera images are resized to $224\times224$ and concatenated into a $224\times448$ input. Visual and action tokens interact through joint attention, with action-loss gradients blocked at the visual-to-action connection (Figure~\ref{fig:delta0_framework}, left).

\paragraph{Dense and sparse joint configurations.}
Fast-WAM-Joint-Dense and Fast-WAM-Joint-Sparse retain joint video--action attention and a 32-action prediction horizon. They differ in the temporal sampling of the visual clip. At decision time $t$, the dense configuration samples the current frame and eight future frames at stride $r=4$, whereas the sparse configuration (Sparse8) samples the current frame and four future frames at stride $r=8$. Thus, both reach $t+32$, with nine versus five sampled frames including the current conditioning frame. The action chunk still contains 32 consecutive commands in both configurations. Table~\ref{tab:fastwam-temporal-layout} specifies the two layouts. Sparse8 refers to the temporal frame stride; the joint-attention architecture is retained.

\begin{table}[!htbp]
\centering
\small
\caption{Temporal layouts of the dense and sparse Fast-WAM-Joint configurations. Offsets are measured in environment steps relative to decision time $t$. Future-frame counts exclude the current conditioning frame.}
\label{tab:fastwam-temporal-layout}
\begin{tabular}{lcc}
\toprule
Setting & Joint-Dense & Joint-Sparse (Sparse8) \\
\midrule
Prediction horizon & 32 steps & 32 steps \\
Action chunk length & 32 commands & 32 commands \\
Visual sampling stride $r$ & 4 & 8 \\
Future frame offsets & $\{4,8,12,16,20,24,28,32\}$ & $\{8,16,24,32\}$ \\
Number of future frames & 8 & 4 \\
Clip frames including the current frame & 9 & 5 \\
Future latent temporal slices & 2 & 1 \\
Future visual tokens & 196 & 98 \\
\bottomrule
\end{tabular}
\end{table}

The sampled clips are encoded with the Wan VAE's causal fourfold temporal compression, yielding two future latent temporal slices for the dense layout and one for Sparse8. With the $224\times448$ two-camera input and the configured tokenization, these correspond to 196 and 98 future visual tokens, respectively; these counts exclude current-frame conditioning and action tokens. The \textit{Gaussian} sparse-joint baseline and our complete \m{}-flow model use the same Sparse8 layout. History-based sources, endpoint-consistency objectives, and action-to-video gradient isolation are specified separately below. The dense-versus-sparse comparison is reported in Section~\ref{sec:discussion-information-coupling}, and measured inference latency is reported in Figure~\ref{fig:inference-latency}.

\paragraph{Sources and objectives.}
The action source uses the previous 32 executed commands in the run's min/max-normalized space, with $\alpha_a=1$ and $\sigma_a=0.5$. Source noise is applied to continuous motion dimensions but not the gripper. The observed visual source clip samples frames at $\{t-32,t-24,t-16,t-8,t\}$, all available at decision time $t$. These frames are encoded and arranged by $\mathbf P_v$ in the future-video latent layout to construct $\mathbf h_v$, and the visual source is $(\mathbf h_v+\bm\epsilon_v)/\sqrt{2}$, a 50\% visual-latent prior with $\alpha_v=\sigma_v=1/\sqrt{2}$ in Eq.~\eqref{eq:difference-source}. Action FM uses visual features at sampled flow times; action EC uses features from the visual source. We set $\lambda_a=\lambda_v=1$ and $\lambda_{\mathrm{EC}}^a=\lambda_{\mathrm{EC}}^v=0.5$, giving
\begin{equation}
\mathcal L_{\mathrm{WAM}}=\mathcal L_{\mathrm{FM}}^a+0.5\mathcal L_{\mathrm{EC}}^a+\mathcal L_{\mathrm{FM}}^v+0.5\mathcal L_{\mathrm{EC}}^v.
\end{equation}
The Gaussian sparse-joint baseline uses the same joint-attention architecture and future-token layout. Comparisons specify matched visual and action sampling budgets; the complete model additionally uses history-based sources, endpoint consistency, and gradient isolation.

\paragraph{Inference and evaluation.}
The complete model uses one visual and one action Euler update, producing a future visual-latent prediction and an action chunk without decoding the visual endpoint. 
 LIBERO-trained checkpoints are evaluated directly on LIBERO and LIBERO-Plus without additional fine-tuning. Shared aggregation and latency procedures are given in Appendix~\ref{app:evaluation-protocols}; training-free source replacement is described in Appendix~\ref{app:training-free}.

\subsection{\texorpdfstring{$\pi_{0.5}$}{pi0.5} Setup}
\label{app:pi05-setup}

\paragraph{Backbone and fine-tuning.}
We initialize $\pi_{0.5}$ from the official \texttt{pi05\_base} checkpoint and implement action \m{}-flow in JAX~\citep{intelligence2025pi05}. Fine-tuning uses \texttt{physical-intelligence/libero}. The main results in Figure~\ref{fig:training-efficiency_and_generalization} use full-parameter fine-tuning. In the additional LoRA experiments, the vision-language Transformer uses rank-16 LoRA with scaling parameter 16, and the action-expert Transformer uses rank-32 LoRA with scaling parameter 32. Non-LoRA parameters in these Transformers are frozen; the visual encoder and action/time projection layers remain trainable. We use continuous state input and supervise seven action dimensions. No visual generation branch is added (Figure~\ref{fig:delta0_framework}, right). Table~\ref{tab:pi05-action-layout} summarizes the action-chunk and inference settings.

\begin{table}[H]
\centering
\small
\caption{Action-chunk and inference settings of $\pi_{0.5}$ with action \m{}-flow.}
\label{tab:pi05-action-layout}
\begin{tabular}{lc}
\toprule
Setting & $\pi_{0.5}$ ($\mathbf{\Delta}$-flow) \\
\midrule
Prediction horizon & 10 steps \\
Action chunk length & 10 commands \\
Action history length & 10 commands \\
Executed actions before replanning & 5 commands \\
Action dimensions & 7 \\
Action sampling steps & 1 \\
\bottomrule
\end{tabular}
\end{table}

\paragraph{Source and objective.}
The action source contains the previous ten executed actions in the checkpoint's quantile-normalized action space. We set $\alpha_a=1$ and $\sigma_a=0.5$; $\mathbf b_a$ excludes gripper and padded dimensions from source noise. The future target is the next ten-action chunk. We set $\lambda_{\mathrm{EC}}^a=0.5$ in Eq.~\eqref{eq:delta-vla-objective}, so the objective is
\begin{equation}
\mathcal L_{\mathrm{VLA}}=\mathcal L_{\mathrm{FM}}^a+0.5\mathcal L_{\mathrm{EC}}^a.
\end{equation}
Action EC uses the L1 discrepancy between the one-step endpoint and target, as defined in Eq.~\eqref{eq:endpoint-consistency}.

\paragraph{Inference and evaluation.}
Action \m{}-flow uses one Euler update and replans after five executed actions. 
We evaluate the model on LIBERO at the end of each epoch.
Full-parameter models are evaluated on LIBERO-Plus. LIBERO-Plus aggregation follows Appendix~\ref{app:evaluation-protocols}.

The Gaussian-source $\pi_{0.5}$ baseline is evaluated with ten-step and one-step Euler sampling.

\section{Evaluation Protocols}
\label{app:evaluation-protocols}

\subsection{Task Success and Aggregation}
Standard LIBERO contains Spatial, Object, Goal, and Long suites.

The four suites organize tasks by their original LIBERO grouping. The seven perturbation
categories organize LIBERO-Plus instances by camera viewpoints, robot initial states,
language instructions, light conditions, background textures, sensor noise, and object layouts.

For LIBERO, Overall is pooled success, $\sum_s S_s/\sum_s N_s$, where $S_s$ and $N_s$ are successes and evaluations in suite $s$. LIBERO-Plus results are reported by perturbation category and task suite, together with Overall success. All breakdowns are obtained from the same evaluation records. Categories follow the official task-index mapping. Published results retain the aggregation used in their original reports.

\subsection{Inference Protocol}
We evaluate the Gaussian-source baselines with ten-step and one-step sampling to compare task success and latency at different inference budgets. The complete WAM variants use one visual and one action Euler update.
The $\pi_{0.5}$ \m{}-flow policy uses one action Euler update. Original ImageWAM and Fast-WAM omit future visual prediction at inference, whereas the joint Fast-WAM variants retain future visual tokens. The complete \m{}-flow models condition action generation on intermediate visual features without decoding future RGB images. Policies execute part of each predicted action chunk before replanning from a new observation. Model-specific chunk lengths and execution horizons follow Appendices~\ref{app:imagewam-setup}, \ref{app:fastwam-setup}, and \ref{app:pi05-setup}.

\subsection{Latency}
The latest measurements use one NVIDIA A100-SXM4-80GB GPU, batch size one, and online text encoding. Timing covers the complete local policy-call pipeline: text encoding, image preprocessing, visual source construction, model prediction, and action postprocessing. This includes ImageWAM image encoding and the historical-video VAE encoding used by complete Fast-WAM. Model loading, compilation, warmup, simulator stepping, rendering, communication, and decoding future RGB images are excluded. 

The WAM implementations use BF16, and the $\pi_{0.5}$ measurements use JAX LoRA models. Solver budgets are compared within each model. The same complete local policy-call timing boundary applies to training-free source comparisons in Appendix~\ref{app:training-free}. Speedup is computed by dividing the reference mean latency by the candidate mean latency.

\subsection{Action Continuity}
The four-way ImageWAM comparison uses Original ImageWAM with ten or one Gaussian action updates,
action-only \m{}-flow with one update, and the complete model with one update.
Trajectories are matched by task and initial state across the four methods.

Action change and action jerk are defined as
\begin{equation}
D_1=\operatorname{mean}_t\|\mathbf a_t-\mathbf a_{t-1}\|_2,
\qquad
D_2=\operatorname{mean}_t\|\mathbf a_t-2\mathbf a_{t-1}+\mathbf a_{t-2}\|_2.
\end{equation}
Boundary jump uses the same first difference only at recorded chunk switches.
Metrics are computed per trajectory and then averaged across trajectories.
Boundary-aligned plots define step zero as the first executed action of the new chunk.

\begin{table}[H] \centering \small \caption{Trajectory-averaged action jerk on ImageWAM. Results are averaged over 20 matched successful trajectories from all ten LIBERO-Long tasks, using the first six action-command dimensions and excluding the gripper. Lower is better.} \label{tab:trajectory-mean-jerk} \begin{tabular}{lcc} \toprule Method & Steps & Mean action jerk $\downarrow$ \\ \midrule \textit{Gaussian} & 10 & 0.04221 \\ \textit{Gaussian} & 1 & 0.19399 \\ \m{}-flow, action only & 1 & 0.04185 \\ \m{}-flow, action + visual & 1 & 0.04335 \\ \bottomrule \end{tabular} \end{table} \paragraph{Trajectory weighting and boundary alignment.}
The trajectory-level metrics give each matched trajectory equal weight,
regardless of episode length. For boundary-aligned curves, we first extract windows from
five control steps before to five steps after each chunk switch. Only boundaries with a
complete window are included. We average the windows within each trajectory and then
average the resulting curves across trajectories. A longer trajectory with more chunk
switches therefore does not automatically receive greater weight in the plotted mean.

\section{Training-Free Source Adaptation}
\label{app:training-free}
\paragraph{Source construction.}
We keep the pretrained policy frozen and modify its inference-time source without updating model parameters. Action-source adaptation uses the executed action history with $\alpha_a=1$ and $\sigma_a=0.5$, with noise restricted to continuous pose dimensions.

\paragraph{Evaluation and timing.}
Comparisons use the same frozen weights, inputs, hardware, and solver budget within each policy. Latency follows the complete local policy-call timing protocol in Appendix~\ref{app:evaluation-protocols}, including source construction.

\section{Action-Objective Training Details}
\label{app:action-objective}
Both configurations use ImageWAM with the $80\%$ image-latent source and retain visual flow matching and visual endpoint consistency. The combined action objective assigns weights of $1.0$ to flow matching and $0.5$ to endpoint consistency. Endpoint-only action training sets these weights to $0.0$ and $1.0$, respectively. Action endpoint consistency uses the L1 discrepancy between the one-step prediction and the future action target. The remaining ImageWAM training settings are given in Appendix~\ref{app:imagewam-setup}.

\section{Action Representation and Control}\label{app:action-representation}
\label{app:action-representation-details}
\paragraph{7D local delta actions.} The action contains three position increments, three axis--angle rotation increments, and one gripper command. The position and rotation increments describe how the end effector should move from its current pose.

\paragraph{7D absolute actions.} The action contains three target-position values, three axis--angle values for the target orientation, and one gripper command. Instead of predicting a pose change, the policy directly specifies the end-effector pose to reach.

\paragraph{10D absolute actions with rotation-6D.} This action specifies the same type of target pose, but uses six values to represent orientation. It contains three target-position values, six rotation values, and one gripper command. The six rotation values encode the first two columns of the target rotation matrix. Before execution, they are converted into a valid rotation matrix by orthonormalization~\citep{zhou2019rotation6d}, then into the absolute pose command accepted by the controller.

\end{document}